\documentclass[lettersize,journal]{IEEEtran}
\usepackage{amsmath,amsfonts}
\usepackage{algorithmic}
\usepackage{algorithm}
\usepackage{array}
\usepackage{amsmath,amssymb,amsfonts}
\usepackage{amssymb}
\usepackage{graphicx}
\usepackage{textcomp}
\usepackage{xcolor}
\usepackage{booktabs}  
\usepackage{threeparttable} 
\usepackage[caption=false,font=normalsize,labelfont=sf,textfont=sf]{subfig}
\usepackage{textcomp}
\usepackage{stfloats}
\usepackage{url}
\usepackage{verbatim}
\usepackage{graphicx}
\usepackage{cite}
\usepackage{hyperref}
\usepackage{dsfont}
\usepackage{verbatim}
\usepackage{multicol}
\usepackage{multirow}
\usepackage{mathrsfs}

\usepackage{xspace}
\makeatletter
\DeclareRobustCommand\onedot{\futurelet\@let@token\@onedot}
\def\@onedot{\ifx\@let@token.\else.\null\fi\xspace}

\def\eg{\emph{e.g}\onedot} 
\def\ie{\emph{i.e}\onedot} 
\def\cf{\emph{c.f}\onedot} 
\def\etc{\emph{etc}\onedot}

\makeatother

\begin{document}

\title{rpcPRF: Generalizable MPI Neural Radiance Field for Satellite Camera}

\author{Tongtong Zhang, Yuanxiang Li*, ~\IEEEmembership{Member, ~IEEE} \thanks{Tongtong Zhang, Yuanxiang Li are with the Department of Information and Control, School of Aeronautics and Astronautics, Shanghai Jiao Tong University.}}



\maketitle

\begin{abstract}
Novel view synthesis of satellite images holds a wide range of practical applications. While recent advances in the Neural Radiance Field  have predominantly targeted pin-hole cameras, and models for satellite cameras often demand sufficient input views. This paper presents rpcPRF, a Multiplane Images (MPI) based Planar neural Radiance Field for Rational Polynomial Camera (RPC). Unlike coordinate-based neural radiance fields in need of sufficient views of one scene, our model is applicable to single or few inputs and performs well on images from unseen scenes.
 To enable generalization across scenes, we propose to use reprojection supervision to induce the predicted MPI to learn the correct geometry between the 3D coordinates and the images. Moreover, we remove the stringent requirement of dense depth supervision from deep multiview-stereo-based methods by introducing rendering techniques of radiance fields. rpcPRF combines the superiority of implicit representations and the advantages of the RPC model, to capture the continuous altitude space while learning the 3D structure.Given an RGB image and its corresponding RPC, the end-to-end model learns to synthesize the novel view with a new RPC and reconstruct the altitude of the scene. When multiple views are provided as inputs, rpcPRF exerts extra supervision provided by the extra views.On the TLC dataset from ZY-3, and the SatMVS3D dataset with urban scenes from WV-3,  rpcPRF outperforms state-of-the-art nerf-based methods by a significant margin in terms of image fidelity, reconstruction accuracy, and efficiency, for both single-view and multiview task.
\end{abstract}
\section{Introduction}
Deep-learning-based large-scale satellite photogrammetry has undergone significant development in both remote sensing and computer vision.  
Novel view synthesis is a computer vision and graphics technique that involves generating new images of a scene or object from viewpoints that were not part of the original data acquisition. 
To efficiently represent the 3D scene,  
neural radiance field (NeRF) has revolutionized the Novel View Synthesis (NVS) task since its introduction by providing more flexible and lightweight solutions.

Obtaining a dense altitude map is another crucial task  that provides reliable and straightforward 3D representations. 
%
Traditional 3D reconstruction methods often rely on multiview or monocular inputs to explicitly recover 3D or 2.5D representations. For satellite photogrammetry, the two pioneering pipelines S2P \cite{s2p} and Sat-COLMAP \cite{satcolmap} are robust but require considerable time for one scene, especially in the dense matching phase. 
In contrast, learning-based multiview stereo (MVS) methods directly perform depth or normal regression \cite{mvsnet, pmnet}, without demanding extensive computations during pairwise matching.
However, computations are still expensive in these approaches. For example, to ensure depth resolution, the regressed cost volume must be cascaded several times \cite{casmvs} or composed in an attention structure \cite{attenmvs,atlasmvs}. In addition, MVS approaches typically require large-scale datasets with dense depth supervision, which may be impractical if the given data are only from optical satellite sensors. Moreover, the mechanism of constructing the cost volumes restricts the model performance on objects with sharp edges such as buildings in urban areas.  
As for the 2.5D representations, such as Layered Depth Images (LDI) \cite{LDI} and Multiplane Images (MPI) \cite{magnifyMPI}, the product  resolutions are often limited by the discrete sampling frequency.
\begin{figure}[t]
    \centering    \includegraphics[width=0.45\textwidth]{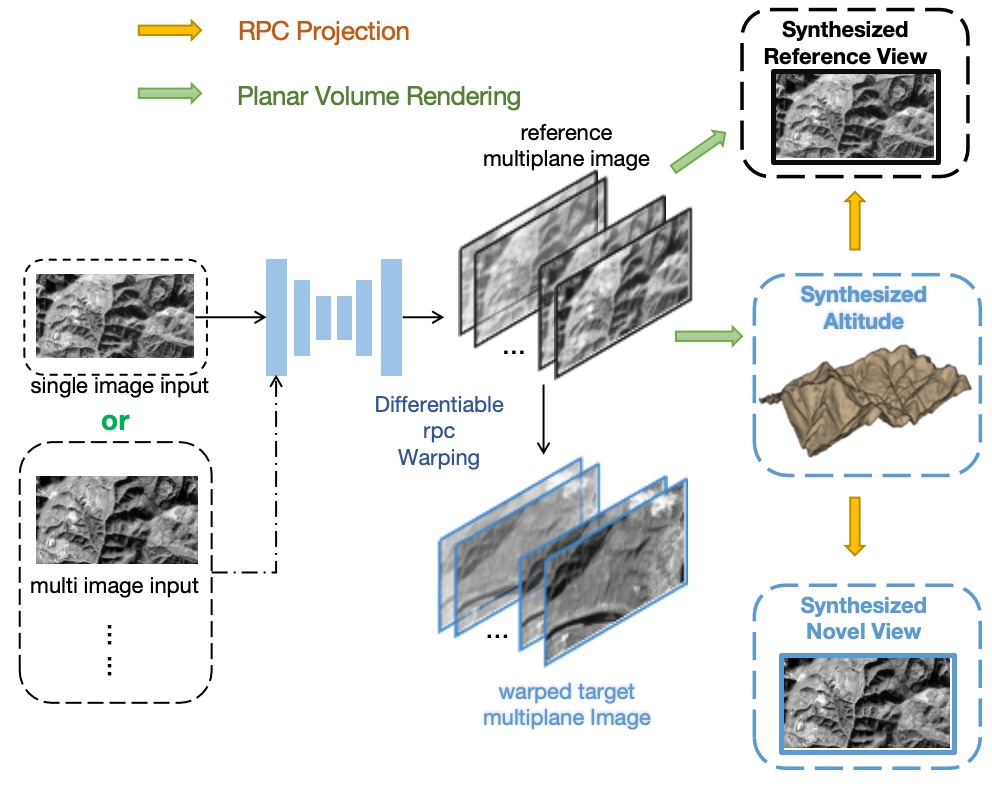}
    \caption{Given a single image or few images from multiple views, based on Rational Polynomial Camera models, rpcPRF produces altitude estimation and synthesize novel views by compositing Multiplane Images with plane volume rendering.}
    \label{fig:intro}
\end{figure}

To circumvent the shortcomings of explicit representations, we resort to the differentiable rendering techniques. 
To date, most successors to NeRF adopt the same idea,  namely a coordinate-based multilayer-perceptron (MLP), to estimate the components per volume for image construction, such as RGB and transmittance. Combining NeRF with other clues gives decent results, such as \cite{neuralmvs,mvsnerf,DSnerf,geonerf,srf}, and the satellite versions \cite{snerf,satnerf} reconstruct a scene without depth supervision.

Nevertheless, there are still two factors that limit the canonical neural radiance field for pushbroom photogrammetry. 
The first factor is the gap between the camera models. 
Although the RPC can be approximated by the pinhole camera, the conversion not only adds time costs, but also introduces unavoidable fitting errors that increase with image size. 
By far, most neural radiance fields are designed for perspective cameras, with the exception of Sat-NeRF \cite{satnerf}.
By changing the sampling strategies, Sat-NeRF can sample rays in geodetic coordinates, the sampling process is one-off,  and it takes a long time to perform the iterative localization step for each position unless caching rays beforehand. 

The second factor of NeRF-based methods is the generalization ability.
NeRF models such as Sat-NeRF and S-NeRF \cite{snerf} are trained and evaluated within one scene, and fail in unseen cases. 
The decoder-based radiance fields are not designed to capture the corresponding geometry between stereo pairs, but only for querying volume information.
However, typical pushbroom sensors have a limited number of shots to keep the photometric properties the same, \eg, the single-line-camera mounted on WorldView-3 (WV-3) takes multiple shots on the same scene within a limited time window. 

Therefore we propose the efficient implicit model rpcPRF, by replacing the coordinated decoder with an encoder-decoder architecture to model a 
planar-density-based field. The convolutional encoder extracts high-level image  features as prior that are concatenated with the embedded altitude to pass to the decoder, and rigorous reprojection losses are exerted as geometric supervision. Instead of rendering per ray, the frustum is sampled in continuous 3D space and warped to the target view by tensor contraction with inverse rational polynomial coefficients. 
Rendering is then performed on each frustum to  estimate the RGB color of the novelty view image, as well as the altitude of the scene.  
As a byproduct of the NVS task, the altitude estimation from the network is based on the inference of the ray termination, according to the RGB cues. Due to the inconsistency of brightness and altitude, the best model for estimating the rendered RGB may not yield the correct altitude map at some typical position, such as buildings cast by shadows and lawns with reflective light, so we add some sparse points as supervision for the altitude estimation task.
%

%
%
In summary, both the single-view and the multiview versions of rpcPRF solve the issues raised by NeRF models, and they
are evaluated on the two satellite multiview stereo benchmark datasets, TLC and Sat-MVS3DM. In the ablation study, we demonstrate the effectiveness of our loss design and the MPI warping module.
The main contributions of this paper are:
\begin{itemize}
    \item It is the first work that applies MPI to pushbroom cameras and uses a planar radiance field to synthesize novel views with new RPC.
    \item rpcPRF is capable of single-view novel view synthesis, and self-supervised altitude estimation with no need for dense ground truth altitude. 
    \item The proposed model generalizes well on unseen scenes, and can be also apply to sparse multiple views.
    \item rpcPRF significantly outperforms existing RPC-based neural rendering models on NVS, while achieving higher inference efficiency, with single or multiple views. 
\end{itemize}

\section{Related work}
\subsection{Deep multiview stereo}
Given a set of images with known camera parameters, MVS aims to recover the dense geometry of the given scene.
Among the learning-based MVS approaches, there are two lines of work, some of which even surpass traditional MVS methods.  
The first line of work  explicitly encodes the stereo pair structure into cost volumes with a learnable image feature encoder and regularizes them with a regression network.
But the feature extraction for 3D cost volume is generally time-consuming and memory-consuming, especially when using 3D convolution \cite{mvsnet}. Some later works have sparsified the cost volume \cite{fastmvs}, and some replace the 3D convolution with a recurrent neural network \cite{rnnmvs} or some attention modules \cite{attenmvs}.
Another line of work adopts a randomized, iterative patchmatch approach to calculate the depth for the nearest neighbors \cite{gipuma, ACMH}, while the end-to-end differentiable version \cite{pmnet} inherits the advantages. The combination of the two ideas is also well-adapted to deep learning settings \cite{pmvsnet}. 
Nonetheless, most of these works require burdensome supervision of dense depth maps, and explicit intermediate results incur huge memory and computation costs.

\subsection{Implicit deep 3D representations}
Implicit Neural Representations (INR) have flourished in these years. Rather than recording the scene as explicit signals, such as depth maps, point clouds, meshes, or voxels, INR tries to learn continuous mappings to represent the signals, which are implemented as neural networks. Common INR paradigms include a coordinate-based decoder that takes coordinate positions as input and returns values at the queried locations,  providing more flexibility for both computation and memory. Surfaces are commonly presented as level sets \cite{deepSDF,deepLS};  scenes as different novel views \cite{srn}, or volumes that can be queried flexibly \cite{reviewnerf,nerf++}.
However, the decoder architecture performs weakly in geometric modeling. Geometry can be incorporated by additionally regularizing the appearance and geometry of patches rendered from unobserved viewpoints, such as 
RegNeRF \cite{regnerf}, or by introducing an encoder. PixelNeRF \cite{pixelnerf}, for instance,
derives a discontinuous scene representation conditioned on the prior encoded by the CNN with only a few images as input. GeoNeRF \cite{geonerf} and MVSNeRF \cite{mvsnerf} leverage the explicit cost volume to encode the geometry.
Most of these methods introduce additional computation while rendering the image per ray, which costs more time and computation than plain NeRF.

\subsection{NeRF for stereo and reconstruction}
Multiview-Stereo and implicit representations complement each other in scene reconstruction.
generally, MVS provides the geometry guidance for the radiance field, such as sparse depth map guidance \cite{nerfingmvs}, or immediate explicit geometrical feature representations as prior \cite{mvsnerf,geonerf}. Structure from Motion (SfM) can also help in exerting depth maps \cite{DSnerf,depthpriornerf} to provide geometry supervision.
Estimating the correspondences based on similar image regions in stereo images can also provide strong supervision of the scene geometry. \cite{srf} learns pair-wise similarities via NeRF, and \cite{mvps} utilize surface normals additionally for confirming the correspondences.

Specific to satellites, S-NeRF \cite{snerf} learns the impact of solar rays while jointly inferring the geometry, and Sat-NeRF \cite{satnerf} first adapts NeRF to pushbroom photogrammetry.
As an extension of S-NeRF, Sat-NeRF 
takes full advantage of metadata of satellite images, by estimating transient objects in addition, similar to NeRF in the wild \cite{nerfw}. However, the models are all designed for one specific scene with sufficient multiple views.
\subsection{Application of MPI}
Owing to the straightforward operations, MPI can be rendered in real-time, with many follow-ups for stereo matching \cite{pushMPI}, novel view synthesis \cite{adaptiveMPI}, video \cite{videompi,magnifyMPI}, \etc.
MPI is typically designed for a stereo pair with a narrow baseline \cite{magnifyMPI}, extracted from the input Plane Sweep Volumes (PSVs) with the estimated alpha constraining the visibility.   
Despite the advantages of MPI in rendering efficiency and quality, the tradeoff between memory usage and the potential rendering ability remains to be a challenge, since the MPI disparity sampling frequency linearly affects  the range of views \cite{pushMPI}.  
For this issue,  \cite{adaptiveMPI} adaptively sparsifies the MPI representations to remove the redundancy by introducing a sparse loss term.
More radically, with the aid of INR, the continuous representation of the discrete MPI extends to the continuous depth space. Nex \cite{nex} combines neural rendering with MPI to fulfill the real-time rendering task; MINE \cite{li2021mine} constructs MPI in the continuous 3D space by introducing the renderer in NeRF; imMPI \cite{immpi} extends MINE by pretraining across scenes and fine-tuning on a single scene, to construct a full light field for aerial images of one scene. 

However, these models can not be directly used on satellite photogrammetry. Except for the discrepancy of the imaging model, the generalization ability across scenes still remains to be settled. 
The most relevant work imMPI \cite{immpi} claimed to render novel views for remote sensing views by modeling the per-scene planar radiance field, but the images come from perspective cameras.
Although the scene priors extracted from cross-scene initialization speed up the per-scene optimization, the model can not generalize to new scenes.

\section{Planar neural radiance field for RPC model}\label{sec:method}
\begin{figure*}
    \centering
    \includegraphics[width=18cm]{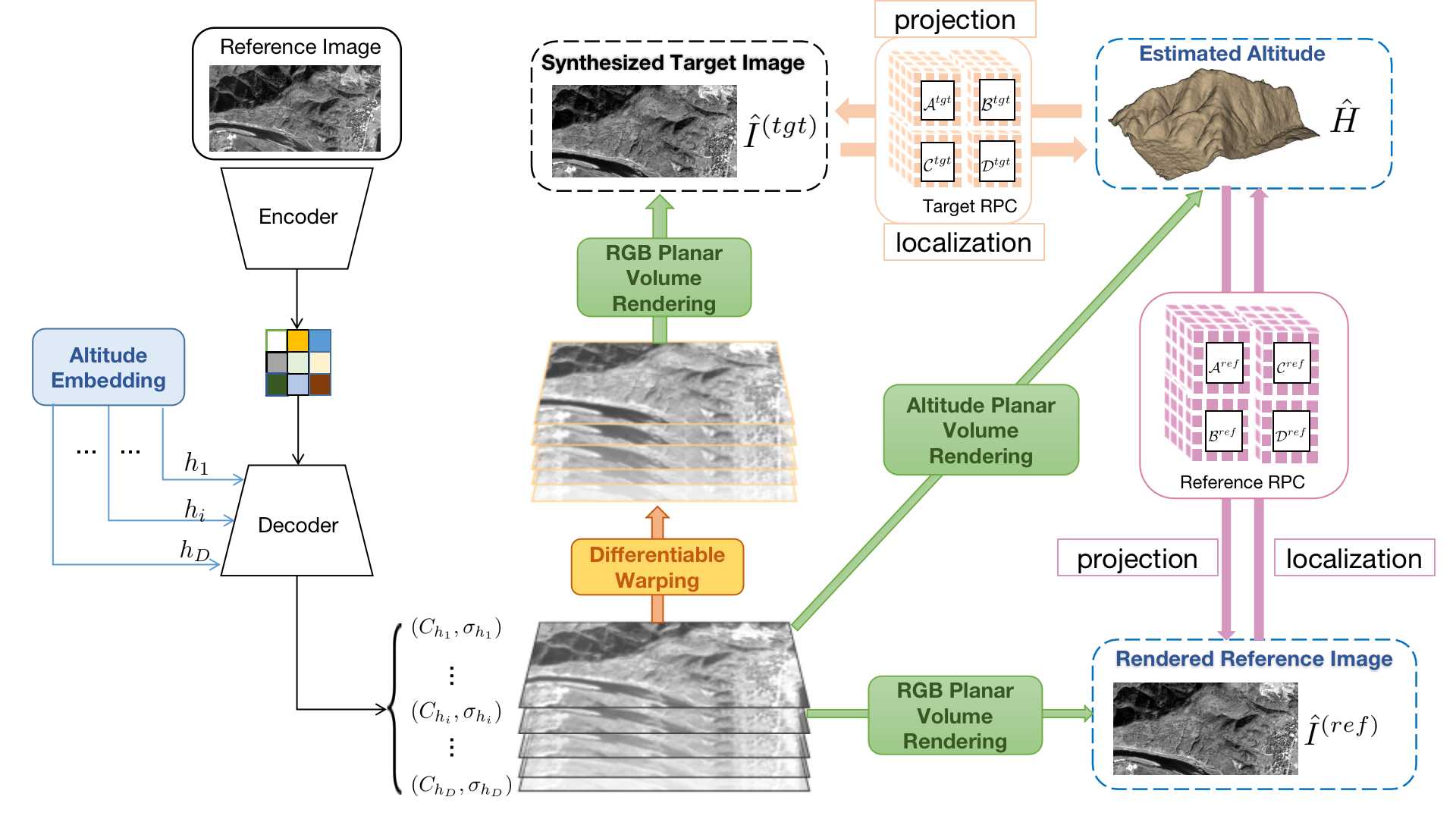}
    \caption{The architecture for the proposed of rpcPRF with a single view, where the photometric supervision comes from the target view $\hat{I}^{tgt}$ and the geometric supervision comes from reprojecting the rendered image back to the rendered altitude $\hat{H}$, \cf Section \ref{sec:single-view}. }
    \label{fig:arch_single}
\end{figure*}

This section firstly elaborates on the proposed rpcPRF with single-view, as the pipeline illustrated in Fig.~\ref{fig:arch_single}. 
Apart from single-view input, rpcPRF with multiple views is described in Section~\ref{sec:multi}. 

\subsection{Efficient implicit 3D representation}

\subsubsection{Conventional MPI}
MPI takes a set of fronto-parallel planes at a fixed range of depths as the basis for global scene synthesis \cite{magnifyMPI}. Different from LDI \cite{LDI}, MPI fixes the layers at sampled depth.

The 3D frustum of the scene is a collection of $N_s$ RGBA layers $\{(C_{z_i}, \alpha_{z_i})\}, i \in \{1, \dots, N_s\}$, where $C_{z_i}(x,y)$ denotes the RGB value of the position $[x, y, z_i]$ with $x,y$ indexing the pixels and $z_i$ indexing the $i^{th}$ sampled depth,
while $\alpha_{z_i}(x,y)$ refers to the alpha map at $[x,y,z_i]$.  
The canonical MPI renderer uses alpha composition on the RGBA set above to yield the image estimation $\hat{I}$ and the disparity map estimation $\hat{D}$:  
\begin{align}
    \left\{
    \begin{array}{cc}
&\hat{I}(x, y)  = \sum\limits_{i=1}^D(C_{z_i}(x, y)\alpha_{z_i}(x, y)\prod\limits_{j=i+1}^D(1-\alpha_{z_j}(x, y)))\\
&\hat{D}(x, y) = \sum\limits_{i=1}^D(z_i^{-1}\alpha_{z_i}(x, y)\prod\limits_{j=i+1}^D(1-\alpha_{z_i}(x, y)))
\end{array}
    \right. \label{eq:alpha_comp}
\end{align}
Then the homography warping is applied to the frustum to render novel views for the perspective cameras.

\subsubsection{Height sampling and embedding}
As metadata determined in advance, the height range $[h_{near}, h_{far}]$ of the region is far larger than common NeRF settings.  To perform proper sampling along the epipolar line, as \cite{invdepth} suggested, we sample $N_s$ height hypotheses between the lowest altitude $h_{far}$ and the highest altitude $h_{near}$ on reciprocal altitude space:
\begin{equation}
\frac{1}{h_i} = \frac{1}{h_{far}} + \frac{i-1}{N_s}(\frac{1}{h_{near}}-\frac{1}{h_{far}}).
\end{equation}

To better capture the comprehensive feature of the large scale height space to initialize the MPI space, we adopt a 1-dim positional embedding to the relative height sample index $h_i, i \in\{0, 1, 2, \ldots, N_s-1\}$ by the frequency encoding:
\begin{align}
       \gamma(h_i) = & [\sin(2^0\pi h_i), \cos(2^0\pi h_i),\ldots, \\ \notag
   & \sin(2^{L-1}\pi h_i), \cos(2^{L-1}\pi h_i)]\label{eq:embed} 
\end{align}
As features in the altitude space, the altitude embeddings are concatenated with multi-scale features extracted with the encoder later.

\subsubsection{Planar radiance field}
The theoretical analysis by \cite{pushMPI} based on signal processing shows that the range of view is limited by the depth sampling frequency. Increasing the sampling frequency helps in representing the stereo, yet the MPI still fails to model the continuous 3D space of arbitrary depth.  

To benefit from the neural rendering techniques for representing a continuous 3D space, MINE \cite{li2021mine} extends the volume rendering of a ray in eq.~\ref{eq:volrender} to planar volume rendering, \ie, for the RGBA set $\{(C_{h_i}, \sigma_{h_i})\}$ at the sampled height set ${h_i}, i\in \{1, 2, \ldots, N_s\}$ for pixel $(x, y)$, $C_{h_i}$ being the color and $\sigma_{h_i}$ being the volume density, the extension of the composition in eq.~\ref{eq:alpha_comp} becomes:
\begin{equation}
 \begin{aligned}
    \left\{
    \begin{array}{cc}
    \hat{I}(x, y) &= \sum\limits_{i=1}^NT_i(x, y)(1-\exp(-\sigma_{h_i}(x, y)\delta_{h_i}))C_{z_i}(x, y)\\
        \hat{H}(x, y) &= \sum\limits_{i=1}^NT_i(x, y)(1-\exp(-\sigma_{h_i}(x, y)\delta_{h_i}))h_i
\end{array}
    \right. \label{eq:volrender}
\end{aligned}   
\end{equation}

where $\delta_{h_i}(x,y) = \|\mathcal{P}(x, y, h_{i+1})^\top-\mathcal{P}(x, y, h_i)^\top\|_2$
denotes the distance between the $i^{th}$ plane to the $(i+1)^{th}$ plane in the pinhole camera coordinate, with  $\mathcal{P}(\cdot)$ being the conversion from perspective 3D coordinate to the Cartesian coordinate. 
$T_i(x,y) = \exp(-\sum\limits_{j=1}^{i-1}\sigma_{h_i}(x,y)\delta_{h_j}), x \in [0, W], y \in [0, H]$ of an image with size $H\times W$
indicates the accumulated transmittance from the first plane $\mathcal{P}(x, y, h_1)$ to the $i^{th}$ plane $\mathcal{P}(x, y, h_i)$, \ie,  the probability of a ray travels from $(x, y, h_1)$ to $(x, y, h_i)$ without hitting the object on the ground surface. 

\subsubsection{Encoder and decoder}
\begin{figure}[t]
    \centering
    \includegraphics[width=9cm]{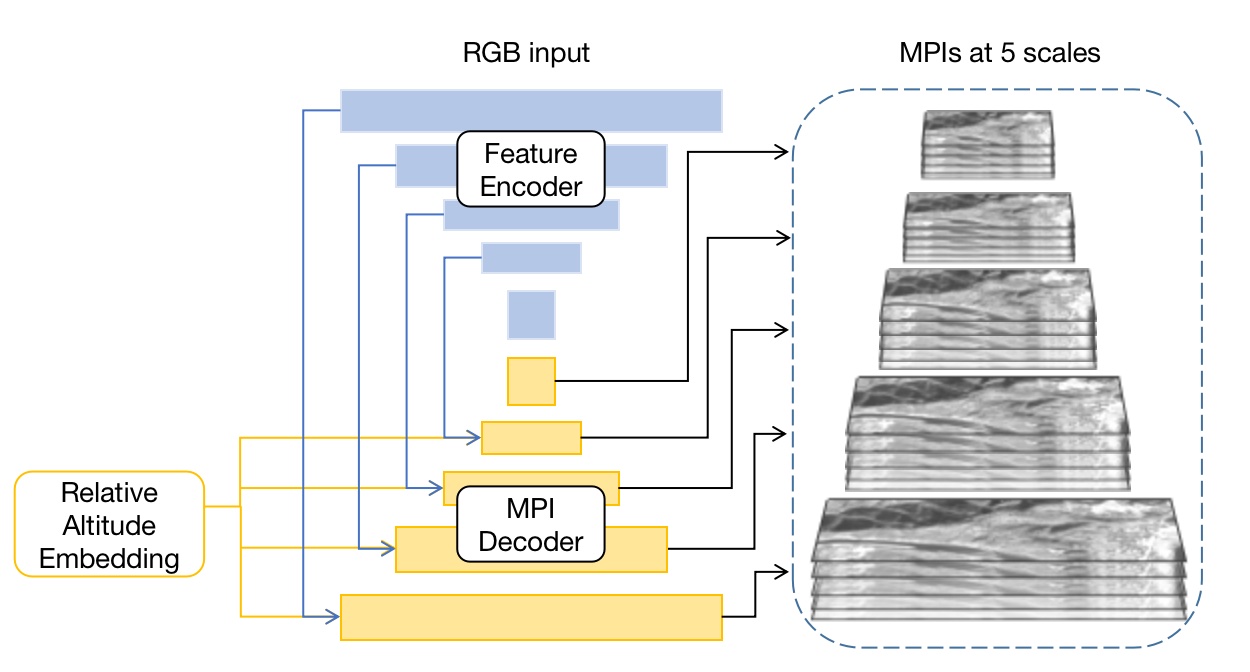}
    \caption{Details of the MPI generator.The image features extracted by the encoder are concatenated with the embedded altitude samples, and fed to the decoder. The MPI is produced by upsample convolution blocks of different sizes.}
    \label{fig:enc_dec}
\end{figure}

To construct the RGBA set $\{(C_{h_i}, \sigma_{h_i})\}, i\in \{1,2,\ldots, D\}$ as the discretized planar radiance field, we adopt an encoder-decoder architecture as \cite{li2021mine} illustrated in Fig.~\ref{fig:enc_dec}. The encoder extracts features from the RGB image and output feature priors at 5 different scales, ${feat_i}, i\in \{1,\ldots, 5\}$, with Resnet-18 as the backbone.
Then the feature set is merged with the embedded heights from eq.~\ref{eq:embed} and fed to the decoder to produce the 4-channel MPI at 5 scales. 
As shown in the factor graph in Fig.~\ref{fig:enc_dec}, the encoder runs one time for RGB images from different views in a batch, and the decoder forwards 5 times for the 5 scales of MPI.

\subsection{Adapting to RPC protocol}
The imaging model of the satellite with RPC serves as an accurate model to project the ground to the corresponding image at pixel level.  
The RPC model offers a compact and precise representation of the geometry between ground and generalized sensors.  It is widely adopted in high resolution satellite photogrammetry, owing to the high approximation accuracy and the independency of the sensor and platform. The RPC model relates the ground coordinate and the image pixel coordinate with the ratios of polynomials. For a 3D point in the normalized world coordinate and its projected 2D point on the image from the sensor, the projection is approximated with polynomials as tensors:

\begin{equation}
    \left\{
    \begin{aligned}
    samp &= \frac{P_1(lat, lon, hei)}{P_2(lat, lon, hei)}=\frac{\mathcal{A}_{(num)}\mathbf{X}}{\mathcal{A}_{(den)}\mathbf{X}}\\
    line &= \frac{P_3(lat, lon, hei)}{P_4(lat, lon, hei)}=\frac{\mathcal{B}_{(num)}\mathbf{X}}{\mathcal{B}_{(den)}\mathbf{X}}\label{eq:rpc_forward}
    \end{aligned}
    \right. 
\end{equation}
where $lat$, $lon$, $hei$ denote the latitude, longitude and height of the 3D point, $samp$ and $line$ are the row and the column of the pixels, and for this point in the object space, its cubic tensor $\mathbf{X} = \{\mathbf{X}_{ijk}\}$ has rank 3, where $\mathbf{X}_{ijk} = [1, hei^i, lat^j, lon^k], i,j,k\in \{0, 1, 2, 3\}$.
And $P_t(\cdot), t\in \{1, 2, 3, 4\}$ represents the cubic polynomials which formulate two ratios to fit the projection from the world to the image. The polynomials are represented by coefficient tensors $\mathcal{A}_{(num)}, \mathcal{A}_{(den)}, \mathcal{B}_{(num)}, \mathcal{B}_{(den)}\in \mathbb{R}^{4\times 4}$ as numerators and denominators respectively. More explicitly, taking $P_1(\cdot)$ for example, there is 
\begin{equation}
    P_1(lat, lon, hei) = \mathcal{A}_{(num)}\mathbf{X} = \sum\limits_{i=0}^{3}\sum\limits_{j=0}^{3}\sum\limits_{k=0}^{3}\mathcal{A}^{ijk}_{(num)}hei^ilat^jlon^k,\label{eq:rpc_coef}
\end{equation}
Inversely, the conversion from image to the ground, \ie localization, can also be approximated by $P_t(\cdot), t \in \{5, 6, 7, 8\}$, where the polynomial for one point can also be represented as tensor contraction, with coefficients on the numerator and denominator as $\mathcal{C}_{(num)}, \mathcal{C}_{(den)}, \mathcal{D}_{(num)}, \mathcal{D}_{(den)}\in \mathbb{R}^{4\times 4}$ respectively,  and its cubic tensor $\mathbf{Y} = \{\mathbf{Y}_{ilm}\}$, with $\mathbf{Y}_{ilm} = [1, hei^i, samp^l, line^m], i,l,m\in \{0, 1, 2, 3\}$ denotes the point in the image space:
\begin{equation}
    \left\{
    \begin{aligned}
    lat &= \frac{P_5(samp, line, hei)}{P_6(samp, line, hei)}=\frac{\mathcal{C}_{(num)}\mathbf{Y}}{\mathcal{C}_{(den)}\mathbf{Y}}\\
    lon &= \frac{P_7(samp, line, hei)}{P_8(samp, line, hei)}=\frac{\mathcal{D}_{(num)}\mathbf{Y}}{\mathcal{D}_{(den)}\mathbf{Y}}
    \end{aligned}\label{eq:inv_rpc}
    \right. 
\end{equation}
With the two-way conversion of one ground point and its corresponding image pixel established, we try to establish the conversion from the reference MPI to the target MPI as batches of points. 

\subsubsection{Differentiable RPC warping in advance}
In Conventional MPI settings, the novel view is rendered via eq.~\ref{eq:volrender} with the MPI of the warped reference view via a homography tensor in the camera coordinate. With the epipolar geometry in homogeneous space, the homography tensor is computed in batches for the source view by composing the intrinsic and extrinsic features of the pinhole camera. 
The homography tensor maps from the reference camera coordinates to the object space and then projects the 3D points onto the source view with a compact tensor operation. Inspired by \cite{satmvs}, 
we also warp the MPI from the reference view to the source views via projection and localization computed in the quaternion cubic homogeneous polynomial space.

\begin{figure}[t]
    \centering
    \includegraphics[width=0.5\textwidth]{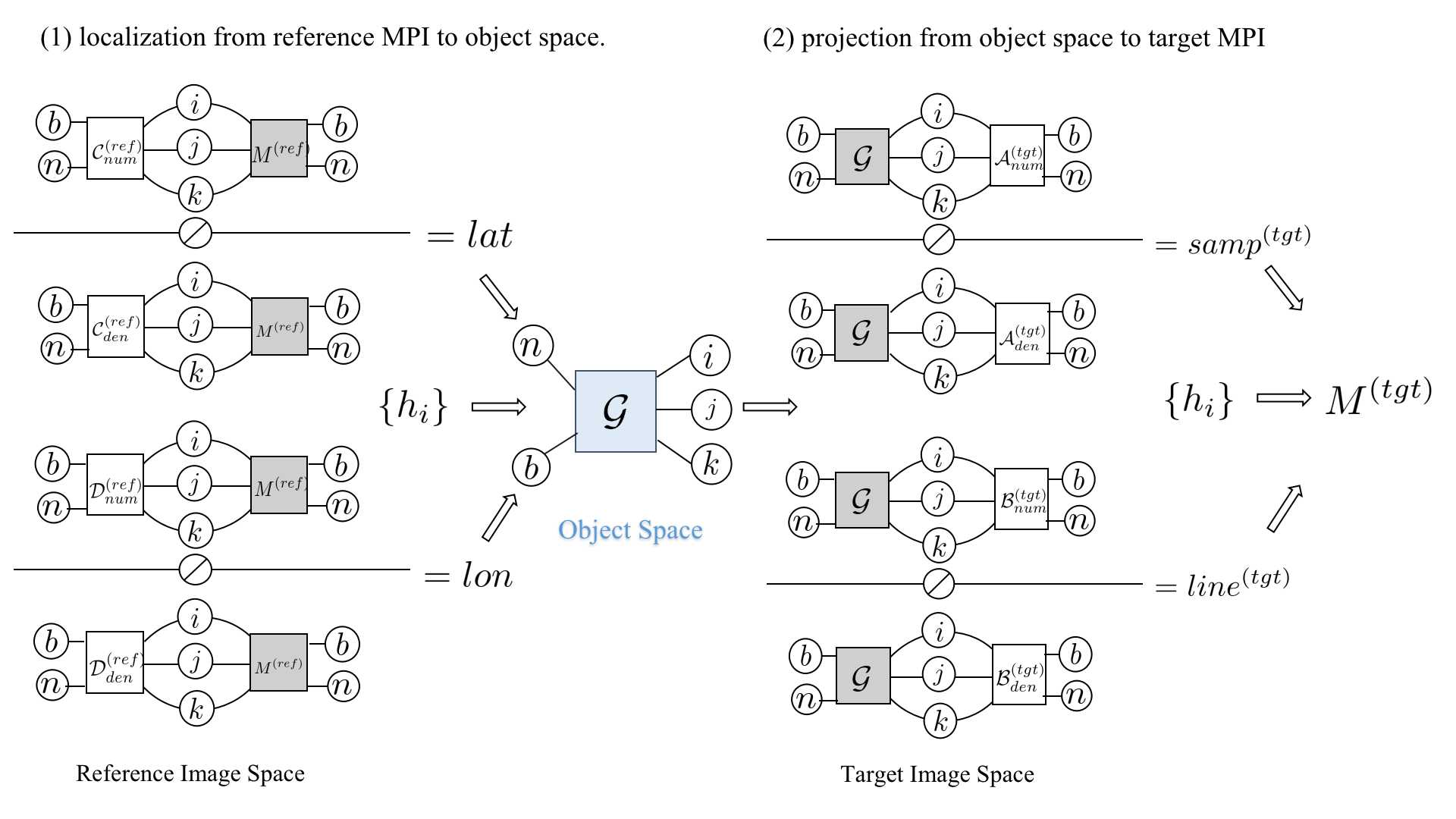}
    \caption{The factor graph of RPC tensor contractions in the batchified conversion of MPI pixels from reference view to target view.}
    \label{fig:factorgraph}
\end{figure}

In the following, we set up MPI warping for RPC 
in the the Geodetic Coordinate System (GCS). Again the forward and inverse mappings as tensors 
in eq.~\ref{eq:rpc_forward} and eq.~\ref{eq:inv_rpc} become tensors of more ranks, allowing for batched computation of points. 
The MPI warping from the reference to the target takes two steps. Firstly virtual grids in GCS are obtained via localization with the reference RPC tensors, and secondly we project the virtual grids to the target MPI via the target RPC tensors. 
Let $B$ denotes the batch size, $N_{mpi}$ denotes the number of points in one set of MPI, for the $n^{th}$ point $M^{bn}_{ilm} = \{(1, (hei^{bn})^i, (samp^{bn})^l, (line^{bn})^m)\}, i, l, m\in \{0, 1,2,3\}$ in the $b^{th}$ batch of MPI points in the image space as a tensor of rank 5 $M^{(ref)} = \{M^{bn}_{ilm}\}$, the tensor composed of the corresponding points in the object space is denoted as $\mathcal{G}=\{\mathcal{G}^{bn}_{ijk}\}$ with $\mathcal{G}^{bn}_{ijk} = \{1, (hei^{bn})^i, (lat^{(bn)})^j, (lon^{(bn)})^k\}, i, j, k\in \{0, 1,2,3\}$, and $b\in \{1,\ldots, B\}, n\in\{1, \dots, N_{mpi}\}$. The tensor index notation of the target MPI in the image space $M^{(tgt)}$ is similar to $M^{(ref)}$, as shown in Fig.~\ref{fig:factorgraph}, where $\oslash$ denotes the element-wise division of tensors.

%
In the first step, the localization process is implemented as contraction of RPC coefficient tensor $\mathcal{C}_{(num)}, \mathcal{C}_{(den)}, \mathcal{D}_{(num)}, \mathcal{D}_{(den)} \in \mathbb{R}^{B\times N_m\times 4 \times 4 \times 4}$ and 
 the MPI tensor $M^{(ref)}\in \mathbb{R}^{B\times N_m\times 4 \times 4 \times 4}$ in the reference image space. 
For the latitude tensor $\mathbf{lat} \in \mathbb{R}^{B\times N_{mpi}}$ and longitude tensor $\mathbf{lon}\in \mathbb{R}^{B\times N_{mpi}}$ respectively, the localization process in (1) of Fig.~\ref{fig:factorgraph} written with tensor notations can be given as: 


 \begin{align}
    \mathbf{lat} &= \mathcal{C}_{(num)}M^{(ref)} \oslash\mathcal{C}_{(den)}M^{(ref)}
      \triangleq \mathcal{F}_{loc}(M^{(ref)},RPC^{(ref)})\\ \notag
    \mathbf{lon} & = \mathcal{D}_{(num)}M^{(ref)}\oslash\mathcal{D}_{(den)}M^{(ref)}\triangleq\mathcal{F}_{loc}(M^{(ref)},RPC^{(ref)}) \label{eq:ten_loc}
\end{align}

where $\mathcal{F}_{loc}$ is defined as the tensor localization function which suits fine to parallel computation with CUDA.

Then the tensor $\mathcal{G}$ of correspondent 3D points in the object space is given by concatenation of the latitude tensor and longitude tensor as $\mathcal{G}=\{hei|samp|line\}$, then we extend the points to its cubic form by combining them with different exponents 
 as $\mathcal{G}^{bn}_{ijk} = \{1, (hei^{bn})^i, (lat^{(bn)})^j, (lon^{(bn)})^k\}, i, j, k\in \{0, 1,2,3\}$, for projection to the target image space with the target RPC. The projection from object space to the target MPI in the second step with tensor notations becomes:


\begin{align}
    \mathbf{samp}  = \mathcal{A}_{(num)}\mathcal{G} \oslash  \mathcal{A}_{(den)}\mathcal{G}\triangleq\mathcal{F}_{proj}(\mathcal{G},RPC^{(tgt)})\\ \notag
    \mathbf{line} = \mathcal{B}_{(num)}\mathcal{G} \oslash  \mathcal{B}_{(den)}\mathcal{G}\triangleq \mathcal{F}_{proj}(\mathcal{G},RPC^{(tgt)})
  \label{eq:ten_proj}
\end{align}

where $\mathcal{F}_{proj}$ is defined as the tensor projection function. 
Therefore, the representations above not only support the two-way conversion in batch, but the coefficients can also be prepared in advance of the network training or inference process, further reducing time and memory consumption,  especially in the localization step. 
The range of altitude in both TLC and SatMVS3DM can be easily queried. Since we observe a much smaller altitude range than that calculated from the RPC, the altitude sample space and the virtual grid from which we pre-compute the RPC tensors are based on the former range.

\subsubsection{Warping for novel view}
\label{warp_frustum}
With the inverse RPC tensor calculated beforehand, and the MPI:
\begin{align}
 & \{(C_{h_i}(samp^{(ref)}, line^{(ref)}), \\ \notag& \sigma_{h_i}(samp^{(ref)}, line^{(ref)}))\}, i\in \{1,\ldots,N_s\}  
\end{align}
 from the decoder for the reference view, for every batch $b$ we get a set of  3D points of size $N_{mpi} = N_s \times H_s \times W_s$ as reference MPI, so we have the tensor representation $M^{(ref)} \in \mathbb{R}^{B\times N_s\times H_s \times W_s}$.
 where $H_s$ and $W_s$ are the height and width of the MPI at scale $s$, and $N_s$ is the number of samplings in the altitude space. 

\begin{figure}[H]
    \centering
    \includegraphics[width=0.5\textwidth]{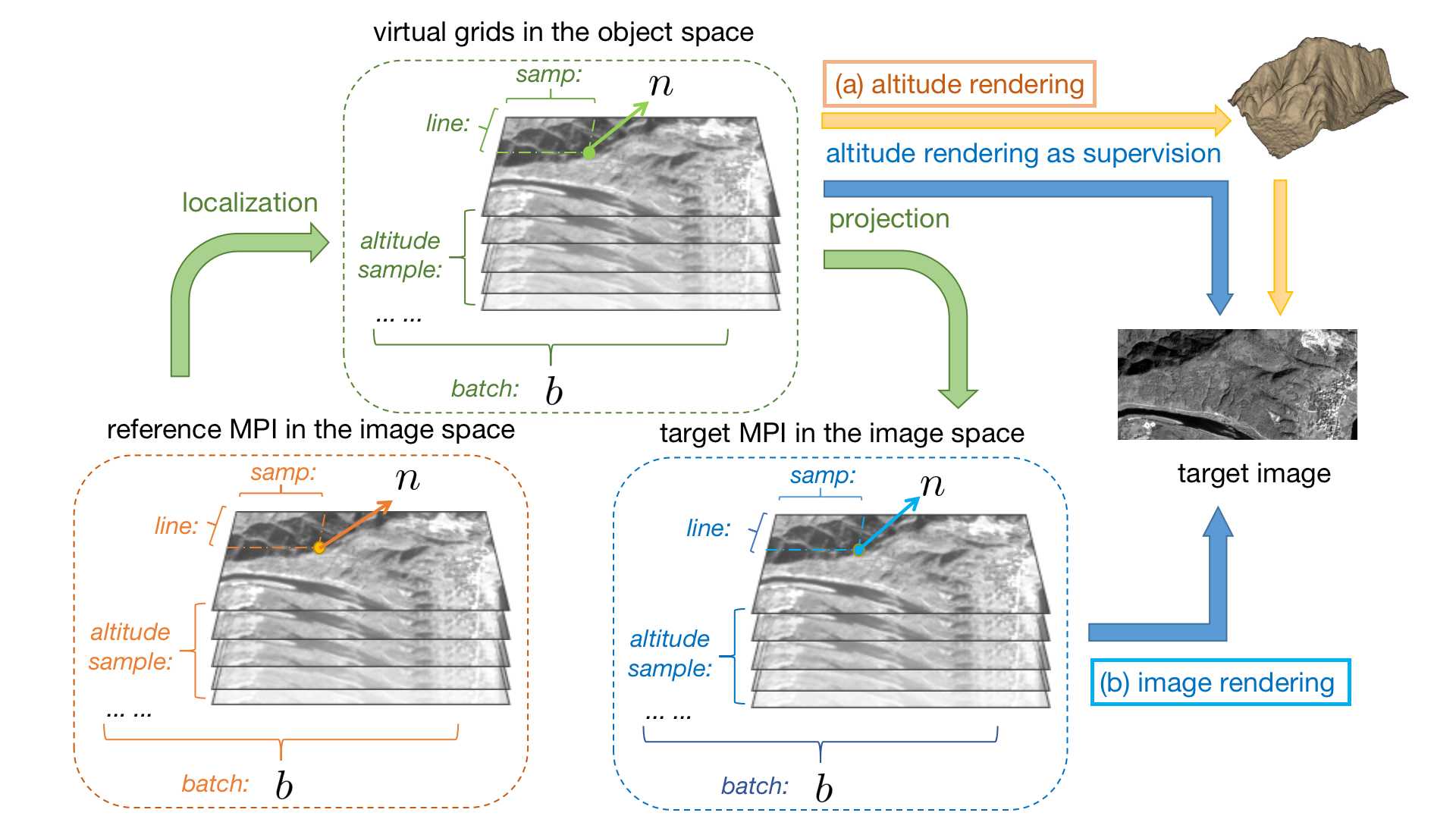}
    \caption{Differentiable warping of the frustum via pre-computed inverse RPC tensors.}
    \label{fig:warpMPI}
\end{figure}

Now there are two ways to obtain the novel view. The instinct one is to use $\mathcal{F}_{proj}$ on the intermediate altitude estimation, as shown in Fig.~\ref{fig:warpMPI} (a). We can obtain the altitude estimation $\hat{H}^{(tmp)}=\{\hat{h}^{(tmp)}\}$ with eq.~\ref{eq:volrender}, and use it as a temporary ground truth height map. 
Using the tensor localization equation with the precalculated inverse coefficients $\mathcal{C}^{(ref)}, \mathcal{D}^{(ref)}$ of reference RPC in \ref{eq:ten_loc}, 
we have the corresponding tensor $\mathcal{G}$ of points in the object space.
However, relying too much on the intermediate results impairs reliability, as intermediate altitude estimates may be imprecise due to unavoidable factors such as shadowing, reflectivity, \etc.

The second way is to warp the whole frustum, and then apply planar volume rendering to the warped MPI in Fig.~\ref{fig:warpMPI} (b). The reference MPI is treated as a batch of points in the image space as a tensor $M^{(ref)}$, with batch size to be the sampling number $D$, the two-step warping process follows Fig.~\ref{fig:factorgraph} to produce $M^{(tgt)}$. 
 Then we obtain the RGB and height estimation via eq. \ref{eq:volrender} with $M^{(tgt)}$.

Despite the convenience of the first way, we validate in the experiments the superiority of the second way, which demonstrates the effectiveness of differentiable MPI warping in both novel view synthesis and altitude estimation.  
\subsection{Loss design}

\subsubsection{Loss functions}
We mainly use two classes of loss functions as supervision according to their functionality, namely, image quality supervision and geometric supervision.
\paragraph{\textbf{Image Quality Supervision}}
Given only the image supervision $I^{(ref)}_{gt}$ and $I^{(tgt)}_{gt}$ as the ground-truth RGB image of the reference and target view, of size $H\times W$ with 3 channel, to induce better render quality for the model, we take $L_1$ RGB loss:
\begin{align}
    \mathcal{L}_{RGB}^{ref} &= \sum\|\hat{I}^{(ref)}-I_{gt}^{(ref)}\|, \\
    \mathcal{L}_{SSIM}^{ref} &= 1-SSIM(\hat{I}^{(ref)},I_{gt}^{(ref)})\label{eq:loss_rgb}
\end{align}
and the structural loss based on the structural similarity index:
\begin{align}
    \mathcal{L}_{RGB}^{tgt} &= \sum\|\hat{I}^{(tgt)}-I_{gt}^{(tgt)}\|,\\
    \mathcal{L}_{SSIM}^{tgt} &= 1-SSIM(\hat{I}^{(tgt)},I_{gt}^{(tgt)})\label{eq:loss_ssim}
\end{align}
\paragraph{\textbf{Geometric Supervision}}
We adopt two kinds of reprojection errors as geometric supervision.
After the height of the reference view $\hat{H}$ is rendered, the first loss project it back to the reference view as $\Tilde{I}_{reproj}=\mathcal{F}_{proj}(\hat{H}, rpc^{(ref)})$ and get the $L_1$ RGB loss:
\begin{equation}
    \mathcal{L}_{reproj}^{ref} = \|\Tilde{I}_{reproj}-I^{(ref)}_{gt}\|\label{eq:loss_ss}
\end{equation}
the other reprojection loss project the points back to the target view as $\Tilde{I}_{reproj}=\mathcal{F}_{proj}(\hat{H}, rpc^{(tgt)})$ and yield another $L_1$ RGB loss:
\begin{equation}
    \mathcal{L}_{reproj}^{tgt} = \|\Tilde{I}_{reproj}-I^{(tgt)}_{gt}\|\label{eq:loss_st}
\end{equation}
\paragraph{\textbf{Sparse Altitude Supervision}}
Optionally, for the altitude estimation task, instead of penalizing the model weights with dense ground truth altitude, we apply sparse ground truth 3D points to facilitate the altitude estimation.  
The models with image quality losses and geometric losses can yield the best estimation of the probability density of ray terminations, based on the hue clues. But the hue clues can be misleading when the colors are brighter at a low altitude, which commonly appear in photos over cities. 
Moreover, sparse supervision is much easier to obtain than dense supervision, with SfM or specified  control points.
For single-view tasks, only the reference view altitude supervision is used:
\begin{equation}
    \mathcal{L}^{ref}_{pts} = \frac{1}{|\#P^{ref}|} \sum\limits_{(r, c, a)\in P^{ref}}(\ln \frac{1}{\hat{H}_{ref}(r,c)}-\ln\frac{1}{a})\label{eq:loss_alt_ref}
\end{equation}
For multiview tasks, the target view altitude supervision can be applied optionally:
\begin{equation}
    \mathcal{L}^{tgt}_{pts} = \frac{1}{|\#P^{tgt}|} \sum\limits_{(r, c, a)\in P^{tgt}}(\ln \frac{1}{\hat{H}_{tgt}(r,c)}-\ln\frac{1}{a})\label{eq:loss_alt_tgt}
\end{equation}
where $\#P^{ref}$ and $\#P^{tgt}$ indicate the number of used 3D points of the reference view and the target view, $\hat{H}_{ref}$ and $\hat{H}_{tgt}$ denote the synthesized altitude of the reference view and the target view, and $r, c, a$ index row, column and altitude of the height map repectively. 
\subsubsection{The total loss}
\label{sec:single-view}
For single-view input, on the novel view synthesis task and height estimation task, we take merely the RGB image and the corresponding RPC of the reference view as input, and render the height estimation $\hat{H}$, reference view image $\hat{I}^{(ref)}$, and the novel view image $\hat{I}^{tgt}$ from the model. The overall loss is given by:
\begin{equation}
    \mathcal{L}_{total} = \lambda_{RGB}\mathcal{L}_{RGB}^{ref} + \lambda_{SSIM}\mathcal{L}_{SSIM}^{ref} + \lambda_{reproj}\mathcal{L}_{reproj}^{ref}\label{eq:loss_single}
\end{equation}
where $\lambda_{RGB}, \lambda_{SSIM}, \lambda_{reproj}$ are the coefficients for the loss respectively.
For altitude estimation task, we further impose sparse points as supervision:
\begin{align}
    \mathcal{L}_{total} &= \lambda_{RGB}\mathcal{L}_{RGB}^{ref} + \lambda_{SSIM}\mathcal{L}_{SSIM}^{ref} \\ \notag
    & + \lambda_{reproj}\mathcal{L}_{reproj}^{ref} + \lambda_{pts}\mathcal{L}_{pts}^{ref}
    \label{eq:loss_single_pts}
\end{align}
where $\lambda_{pts}$ is the weight for points loss.
\subsection{rpcPRF for multiple views}
\label{sec:multi}
\subsubsection{Pipeline}
rpcPRF is not limited to a single view if there are more images provided.  Rather than forwarding the encoder and decoder many times, multiple views provide stronger supervision on target view synthesis without increasing total parameters. The difference between rpcPRF on single-view and multiple views are shown in Fig.~\ref{fig:pipeline_diff}:
\begin{figure*}
    \centering
    \includegraphics[width=18cm]{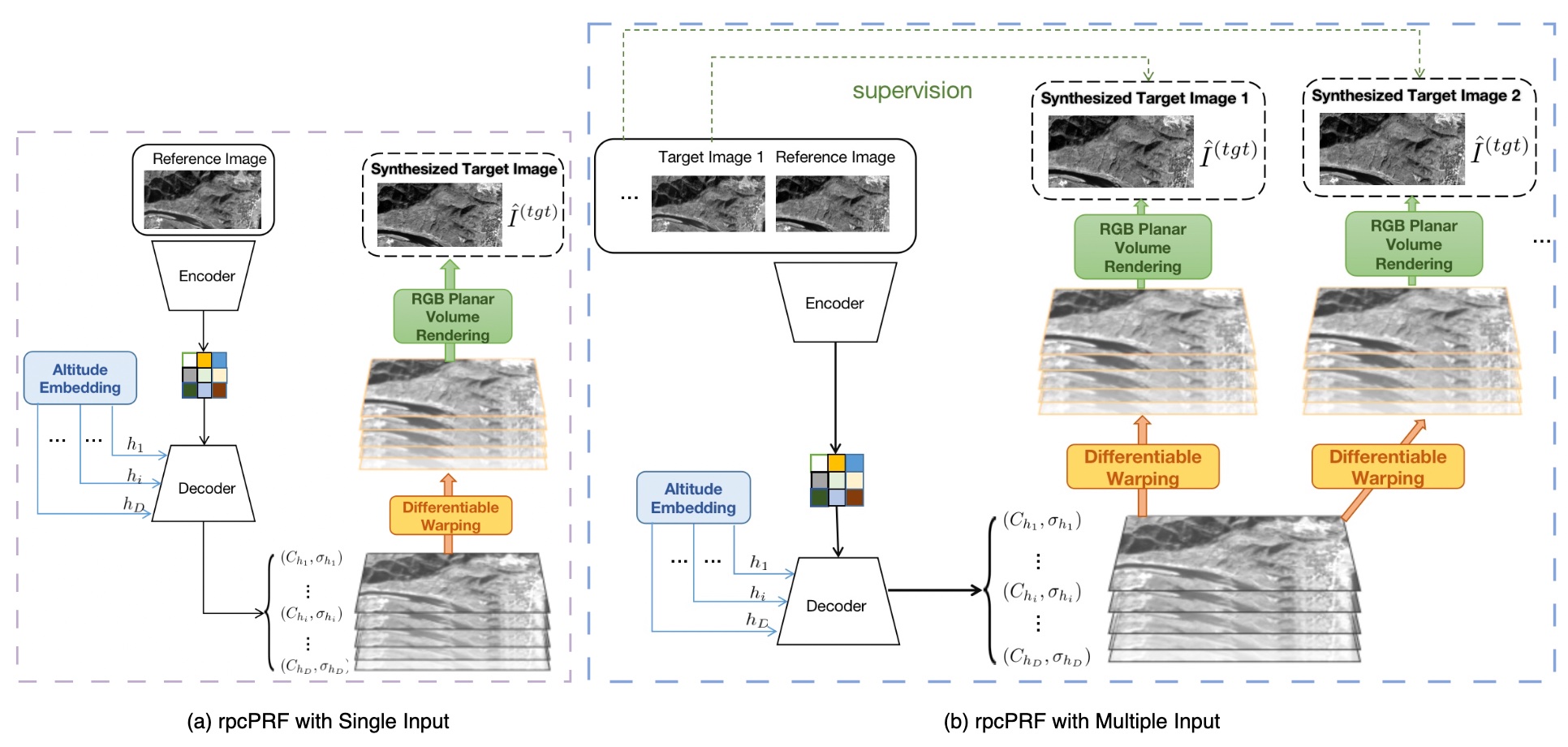}
    \caption{The difference of rpcPRF pipelines with single-view and multiple views.}
    \label{fig:pipeline_diff}
\end{figure*}
\subsubsection{The total loss}
\label{sec:multiview}
For multiview input, on the tasks novel view synthesis and height estimation, the input consists of the RGB images of the reference and the target views, and the loss consists of the penalizations on the reference view and the target view:
\begin{equation}
   \begin{aligned}
    \mathcal{L}_{total-m} &= \lambda_{RGB}(\mathcal{L}_{RGB}^{ref} + \mathcal{L}_{RGB}^{tgt})\\ \notag
    & + \lambda_{SSIM}(\mathcal{L}_{SSIM}^{ref} 
    + \mathcal{L}_{SSIM}^{tgt})   
    \label{eq:loss_mvs}
\end{aligned} 
\end{equation}


\section{Experimental results and analysis}
This section presents quantitative and qualitative evaluations of our methods on the two representative datasets for RPC-based satellite photogrammetry, TLC \cite{satmvs} and SatMVS3D \cite{lujiacheng}.  First, we introduce the details of the datasets  and our preprocessing. Second, we present model setups for view synthesis and altitude estimation on single-view tasks and multiview tasks.  Then we explicitly specify the details of the comparison with the state-of-the-art methods on RPC photogrammetry, together with the ablation studies.  Finally, we conduct ablation studies on the proposed MPI warping module and losses, and then we analyze how the hyperparameters affect the models. 
\subsection{Data preparation}
\paragraph{\textbf{TLC}}
The TLC SatMVS dataset are collected from the triple-line-camera mounted on ZY-3 satellite \footnote{\url{http://gpcv.whu.edu.cn/data/whu_tlc.html}}. The optical images are naturally organized as triples of 2.1 m in resolution for the nadir view and 2.5 m in resolution for the two side views.
The original TLC dataset provides ready-made splits for training and testing, from which we select parts of the dataset, and further split the original training sets into a training set of 800 triples and a validation set of 100 triples, the test set are of 200 triples. All of the ground truth RGB images and the height maps are composed of 384 x 768 pixels.
In addition, we remove the NO-DATA value of the ground truth height map to the minimum height of the valid value, which takes a large proportion. We then recalculate the RPC with the numerator and denominator for samp and line in localization. The virtual grid for this precomputation shrinks to the altitude range with a larger latitude and longitude range.
For single-view novel view synthesis and altitude estimation, the triple are reduced to the nadir view. And for the multiview task we select the nadir as the reference view and the other two as the target views.
\paragraph{\textbf{SatMVS3DM}}
The SatMVS3DM dataset is an adaptation of the original Multi-View Stereo 3D Mapping Challenge (MVS3DM) challenge dataset for the MVS task.  Originally the panchromatic and multispectral images from MVS3DM challenge \footnote{\url{https://spacenet.ai/iarpa-multiview-stereo-3d-mapping/}}
 focus on a 100 $km^2$ area near San Fernando, Argentina, collected over a span of two years by the WV-3 satellite with a resolution of 30cm per pixel in nadir views. The SatMVS3D dataset converts the ground truth point cloud from UTM to geodetic coordinates, which are then projected onto image coordinates to obtain the ground truth height map. On the basis of SatMVS3D, we filter out the noise of the height map, which deviates far from the ground surface, hence guaranteeing a legitimate altitude range for sampling.  
For every scene, we take an image size of $384\times 480$. As a supplementary of the selected triple-stereo from SatMVS3DM, we randomly generate groups of subsets of each scene with sizes of 5, 7 and 9, which corresponds the notation g3, g5, g7, g9 in Table~\ref{tab:data}.
The original ground truth of 20cm airborne lidar ground truth data for a subset of 20 square kilometers of the same area. There are 6 sites of different sizes within the dataset, the split and the abbreviation information is detailed in Table~\ref{tab:datasite}.
\begin{table}[H]
    \centering
     \caption{Dataset description details of TLC dataset and the site MasterProvisional3 of SatMVS2DM dataset with different group of size N.}
\begin{tabular}{c|ccccc}  
    \toprule 
    
    Dataset Name&\multicolumn{1}{c}{\textbf{TLC}}&\multicolumn{4}{c}{\textbf{MP3}}\cr  
    \cmidrule(lr){3-6} 
    group name& &g3&g5&g7&g9\cr  
    \midrule  
    train& 600 &500 &560 & 512& 470  \cr
    val& 120& 100& 112& 102& 100 \cr
    test& 120& 100&112 &102 &100 \cr
    \bottomrule  
    \end{tabular}   

    \label{tab:data}
\end{table}

\begin{table}[H]
    \centering
     \caption{Splits and abbreviation of different Sites of the SatMVS3DM dataset.}
\begin{tabular}{c|ccc|c}  
    \toprule 
    site name& train&val&test&abbr\cr  
 \midrule
    Explorer& 1564 &312 &120 & Exp  \cr
    MaterProvisional3& 500& 100& 100& MP3 \cr
    MaterSequestered1& 571& 114&60 &MS1 \cr
    MaterSequestered2& 588& 117&30 &MS2 \cr
    MaterSequestered3& 554& 110&90 &MS3 \cr
    MaterSequesteredPark& 401& 80&60 &MSP \cr
    \bottomrule  
    \end{tabular}   
    \label{tab:datasite}
\end{table}

\subsection{Experiments setup}
\subsubsection{Evaluation metrics}
For altitude estimation,
after the conversion of the height map to the DSM, we further quantitatively evaluate the reconstruction quality with common metrics: 
\begin{itemize}
    \item The mean absolute error (\textbf{MAE}): the average of the L1 distance over all the grid units between the ground truth and the estimated height map;
    \item Median Height Error (\textbf{ME}): The median of the absolute L1 distance over all the grid units between the ground truth and the estimated height map;
    \item Error below thresholds: the percentage of grid units with an L1 distance error below the thresholds of 1.0 m (approximately equivariant to the ground sample distance (GSD)), 5.0m and 7.5 m ($<1.0m$, $<2.5m$ and $<7.5m$);
    \item The inference time (\textbf{time}): the inference time from feeding a test image to the network until the height map is inferred.
\end{itemize}
For novel view synthesis, we take the common image quality indicators, the learned perceptual image patch similarity (LPIPs), the peak signal-to-noise ratio (PSNR), and structural similarity index measure (SSIM)  
 as metrics. LPIPS metric utilizes a VGG network to evaluate image similarity at the feature level, while PSNR and SSIM evaluate the image similarity at the pixel level. 


\subsubsection{Training detatils}
We trained the proposed rpcPRF with various combination of losses, and we take a ResNet-18 as encoder, a height decoder with similar structure as Monodepth2 \cite{monodepth2}. The rpcPRF models are implemented with PyTorch, and the compared models NeRF and Sat-NeRF are implemented with Pytorch Lightning. All of the models are trained on an NVIDIA RTX 3090 (24GB) GPU. For rpcPRF, we adopt Adam optimizer with a multi-step scheduler for the learning rate. The initial learning rates for the encoder and decoder are set to be 0.0001 and 0.0002, and the milestones for the schedulers vary according to the number of parameters and the epoch settings. The maximum number of epochs is 100 for the multiview setting and 50 for the single-view setting For comparison fairness, we utilize the own settings of the compared models to ensure their performance, besides we modify the dataloader and ray batch size to apply to our problem. 
\begin{figure*}
    \centering
    \includegraphics[width=18cm]{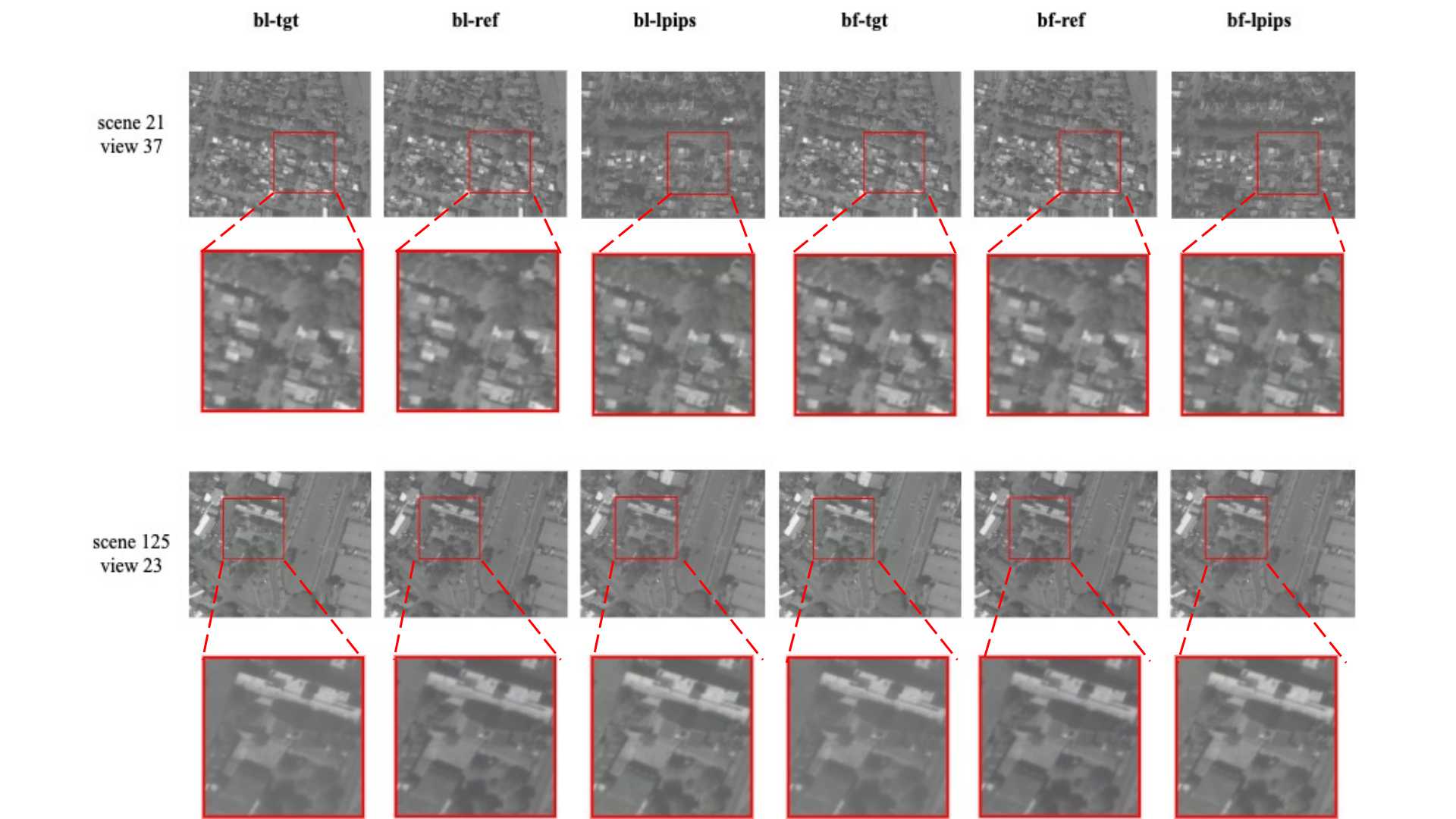}
    \caption{Qualitative comparison of different loss impacts on single-view rpcPRF for novel view synthesis on the MP3 subset of Sat-MVS3DM Dataset. }
    \label{fig:single_nvs}
\end{figure*}

\subsection{Comparison with State-Of-the-Art on novel view synthesis}
\subsubsection{Comparison methods}
To date, the novel view synthesis works based on neural radiance fields have been rarely designed for pushbroom cameras. The derivation of the two representative works, Shadow-NeRF (S-NeRF) \cite{snerf} and Sat-NeRF \cite{satnerf} both require extra meta-information, such as  solar supervision for calculating the irradiance field. Since the two cross-scene multiview datasets we chose have no more information,  we adapt our relative comparison as follows: 
\begin{itemize}
    \item \textbf{NeRF(RPC)}:
    As a corner stone of neural implicit representations, NeRF models the scene volume with MLP networks as a mapping function, which means that the functions to query volumes require per-scene optimization. For the sake of fairness, we train the MLP functions across scenes, which is the basic setting of our proposed model. 
    Both validation and testing are performed on unseen scenes from the training set, with the same ratio as our rpcPRF split shown in Table~\ref{tab:data}, but with the input format as rays and pixels instead of the entire image.
    The model parameters are set the same as in S-NeRF except for the lighting condition modules.
    Moreover, we adopt the ray sampling procedure similar as Sat-NeRF for the most plain NeRF comparison without modeling the shadows.
    \item \textbf{Sat-NeRF}:
    Despite meta-information deficiency, we extend the existing information for NeRF-based models as much as possible, based on RPC ray sampling. Firstly, we preserve the embedding vector $t_j$ learned as the function of the image index $j \in\{1,2,3\}$, to encode the effect caused by the relative position, since there is no more geometry encoded. Secondly, we hold the uncertainty coefficient related to the probability of explaining whether the transient object explains the color.
    Since the features of these two factors are concatenated with the sun direction tensor and then fed to the corresponding network, we set the sun directions for all the input images to be perpendicular to the ground, serving as an empty placeholder.
    The other training settings follow Sat-NeRF, \eg, the stopping rule of training is the inflection point of validation PSNR,  and the dataset splitting follows our model.
    \item \textbf{rpcPRF-s}: The single-view version of rpcPRF is abbreviated as rpcPRF-s, trained with the loss in eq.~\ref{eq:loss_single}. Instead of validation PSNR, the training of rpcPRF-s stops at the inflection point of the total loss.

    \item \textbf{rpcPRF-m}: The multiview version of rpcPRF is abbreviated as rpcPRF-m, trained with the loss in eq.~\ref{eq:loss_mvs}.
\end{itemize}
All the models above are designed for the different triples of images, and our goal is to compare the generalization ability of the models on unseen scenarios.
All the models are trained on the 6 sites of SatMVS3DM dataset, with splits and abbreviation details in Table~\ref{tab:datasite}. All the metrics are computed as averages across each site itself. 

\begin{figure*}
    \centering
    \includegraphics[width=18cm]{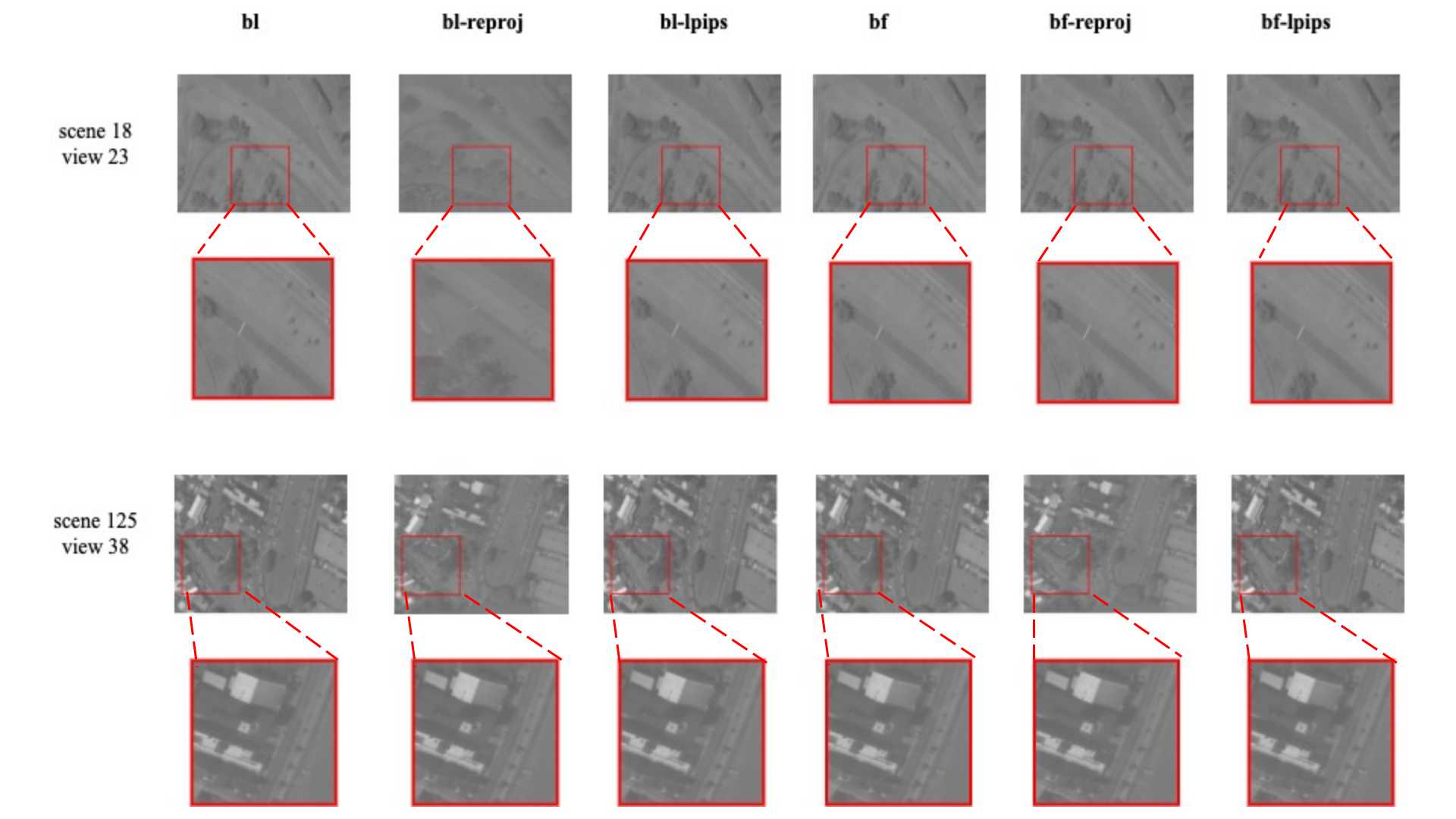}
    \caption{Qualitative comparison of different loss impacts on multiview rpcPRF for novel view synthesis on the MP3 subset of Sat-MVS3DM Dataset. }
    \label{fig:comp_multi_bl}
\end{figure*}

\subsubsection{Image quality comparison and discussion}
\begin{figure*}
    \centering
    \includegraphics[width=0.9\textwidth]{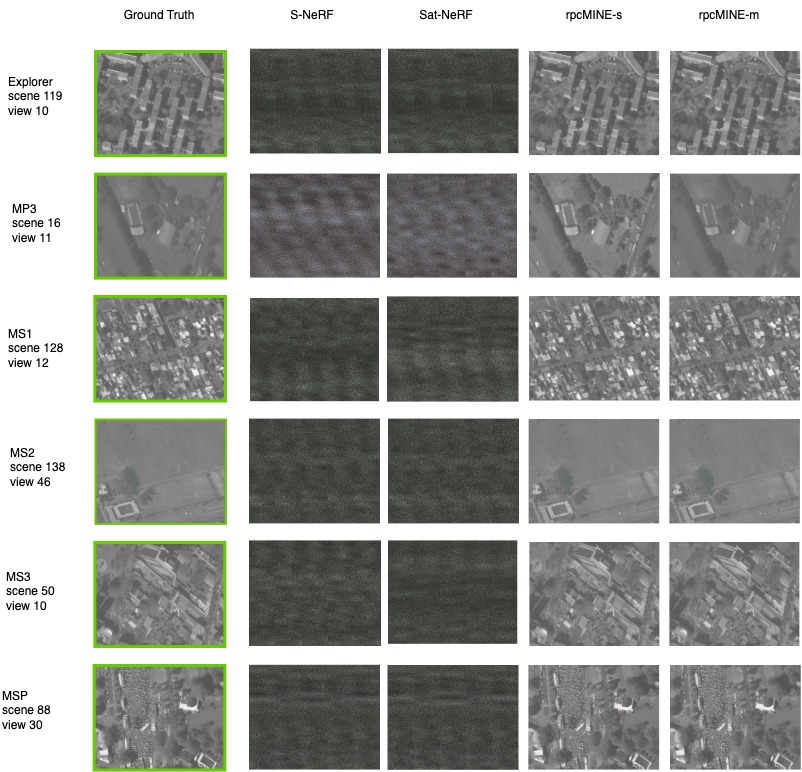}
    \caption{Qualitative comparison of the models on the test triples of the SatMVS3DM dataset. The test triples are all unseen to all the four models, 
    rpcPRF-s is trained with eq.~\ref{eq:loss_single}, rpcPRF-m is trained with eq.~\ref{eq:loss_mvs}}
    \label{fig:comp_nerf_mine_IARPA}.   
\end{figure*}

\begin{figure*}
    \centering
    \includegraphics[width=18cm]{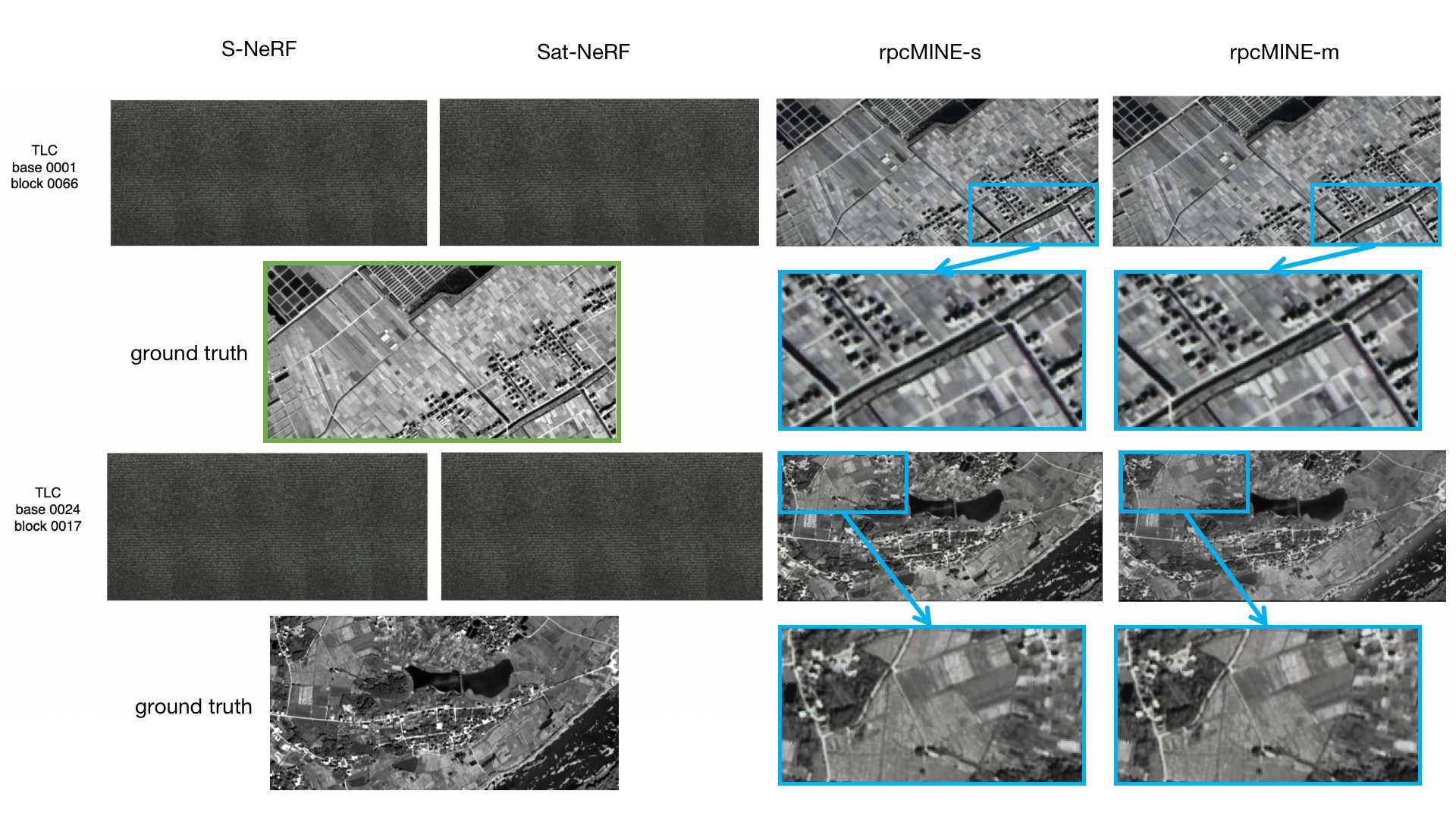}
    \caption{Qualitative comparison of the models on the test triples of the TLC dataset. The test triples are all unseen to all the four models, 
    rpcPRF-s is trained with eq.~\ref{eq:loss_single}, rpcPRF-m is trained with eq.~\ref{eq:loss_mvs}}
    \label{fig:comp_nerf_mine_TLC}.   
\end{figure*}

Table~\ref{tab:comp_nerf_mine} shows the quantitative comparison between rpcPRF and NeRF-based methods on different sites of Sat-MVS3DM, Table~\ref{tab:comp_nerf_mine_TLC} on TLC,  and Fig.~\ref{fig:comp_nerf_mine_IARPA} shows the qualitative comparison between these methods on Sat-MVS3DM, and Fig.~\ref{fig:comp_nerf_mine_TLC} shows the 
 result on TLC dataset.
\begin{table*}
  \centering  
      \caption{Quantitative comparison of test results of state-of-the-art methods on SatMVS3DM dataset.} 
  \fontsize{7}{7}\selectfont 
  \LARGE
  \resizebox{16cm}{!}{
  \begin{threeparttable}  
    \begin{tabular}{ccccccccccccc}  
    \toprule  
    &\multicolumn{4}{c}{\textbf{PSNR}$\uparrow$}&\multicolumn{4}{c}{\textbf{SSIM}$\uparrow$}&\multicolumn{4}{c}{\textbf{LPIPS}$\downarrow$}\cr  
    \cmidrule(lr){2-5} \cmidrule(lr){6-9}\cmidrule(lr){10-13}  
    Sites&NeRF(RPC)&Sat-NeRF&rpcPRF-s&rpcPRF-m&NeRF(RPC)&Sat-NeRF&rpcPRF-s&rpcPRF-m&NeRF(RPC)&Sat-NeRF&rpcPRF-s&rpcPRF-m\cr 
    \midrule  
Explorer&13.072 &13.588&\textbf{26.982}&25.004&0.127&0.195&0.704&0.729&0.803&0.793&0.281&0.388\\
MP3& 12.062&13.608&25.499&26.323&0.092&0.103&0.806&0.801&0.894&0.803&\textbf{0.194}&0.199 \\
MS1& 11.017&11.482 &22.275&23.856&0.081&0.088&0.794&0.624&0.814&0.882&0.299&0.336  \\

MS2&12.872 &12.351&23.784&25.619&0.098&0.096&\textbf{0.809}&0.734&0.758&0.792&0.228&0.265 \\
MS3&10.869 &12.014&23.336&26.811&0.082&0.114&0.725&0.748&0.744&0.698&0.352&0.388\\
MSP&11.320 &11.975&23.554&23.199&0.088&0.090&0.675&0.775&0.832&0.856&0.247& 0.361 \\
\hline
mean &11.830	&12.503&24.419&25.135	&0.0947&0.114 &0.752	&0.735&0.808	&0.804	&0.260	&0.299  \\
    \bottomrule  
    \end{tabular}  
    \end{threeparttable}
    }
    \label{tab:comp_nerf_mine}
\end{table*}
\begin{table*}
  \centering  
    \caption{Quantitative comparison of test results of state-of-the-art methods on TLC dataset.}  
  \fontsize{7}{7}\selectfont
  \LARGE
  \resizebox{16cm}{!}{
  \begin{threeparttable}  
    \begin{tabular}{ccccccccccccc}  
    \toprule  
    &\multicolumn{4}{c}{\textbf{PSNR}$\uparrow$}&\multicolumn{4}{c}{\textbf{SSIM}$\uparrow$}&\multicolumn{4}{c}{\textbf{LPIPS}$\downarrow$}\cr  
    \cmidrule(lr){2-5} \cmidrule(lr){6-9}\cmidrule(lr){10-13}  
    Sites&NeRF(RPC)&Sat-NeRF&rpcPRF-s&rpcPRF-m&NeRF(RPC)&Sat-NeRF&rpcPRF-s&rpcPRF-m&NeRF(RPC)&Sat-NeRF&rpcPRF-s&rpcPRF-m\cr 
    \midrule  
TLC &14.088&14.512&23.778&\textbf{24.473}&0.092&0.098&0.775&\textbf{0.782}&0.892&0.874&0.301&\textbf{0.298}\\
    \bottomrule  
    \end{tabular}  
    \end{threeparttable}
    }
    \label{tab:comp_nerf_mine_TLC}
\end{table*}

As can be seen from Table~\ref{tab:comp_nerf_mine}, Fig.~\ref{fig:comp_nerf_mine_IARPA} and Fig.~\ref{fig:comp_nerf_mine_TLC}, the two NeRF-based methods fail to generalize across scenes, while both rpcPRF-s nd rpcPRF-m take the lead on the three metrics for all the scenes.
The innate design of the former two methods does not encode the geometry, which means that it performs poorly on the reconstruction task. In the original setting for Sat-NeRF \cite{satnerf}, the goal is to learn the volumes to construct structure within a specific scene and render the novel view of the same scene, while rpcPRF aims to learn the geometry of the underlying scene via RPC. Sat-NeRF performs tenuously better than S-NeRF in this case, which shows the importance of image indexing embeddings to some extent, but the indexing embeddings still fail to capture the geometry. In contrast, both the single-view and multiview version perform well on unseen cases. The single-view rpcPRF slightly outperforms PSNR, SSIM and LPIPS.
\subsubsection{Efficiency comparison}
In addition to the image quality assessment, we also compare the models by memory and computation consumption, time cost, and the model parameters. It is important to note that rpcPRF-s and rpcPRF-m differ in inputs and the training process, so the memory and computation consumption is the same at inference time, thus three models are available for comparison in Table~\ref{tab:comp_nerf_memory}:   
\begin{table}[H]
    \centering
     \caption{Time and computation comparisons of NeRF-based methods and the proposed rpcPRF-s and rpcPRF-m}
      \label{tab:comp_nerf_memory}
    \begin{tabular}{c|cccc}
    \toprule
        sites & \textbf{Memory}$\downarrow$ & \textbf{Param}$\downarrow$& \textbf{FLOPs}$\downarrow$& \textbf{Time}$\downarrow$\\
        \midrule
NeRF(rpc)&2.863G & 0.63M& 103.08G& 23.295s \\
Sat-NeRF&3.147G &0.66M &338.43G & 24.263s\\
rpcPRF-s(m)& 0.477G&19.79M&202.23G & 0.954s\\
         \bottomrule
    \end{tabular}
\end{table}
The two metrics \textbf{FLOPs} and \textbf{Time} are calculated for a triple of images. NeRF-based models are rendered per ray, so the temporary memory actually depends on the batch size of rays, while rpcPRF series rendered per image, so we set the batch size of S-NeRF and Sat-NeRF to be $3 \times H \times W$.
Table~\ref{tab:comp_nerf_memory} shows that the proposed rpcPRF takes less computation than Sat-NeRF, with faster inference process. 
Besides, for most of current single-gpu settings, the batch size of $3\times H\times W$ rays is unattainable, so the total inference time of rpcPRF is actually far less than NeRF-based models.

\subsection{Single-view rpcPRF and ablation study}
This section explores the impacts of the key factors on rpcPRF. The first consideration is to validate the effectiveness of the differentiable frustum warping module for novel view synthesis, and we take the novel view as a direct projection from the estimated altitude for comparison \ref{sec:method}. 
The second consideration is to validate the impact of different losses combinations on the single-view rpcPRF for the two tasks.
\subsubsection{Model notations}
 The baseline notations for novel view synthesis are listed below:
\begin{itemize}
    \item \textbf{bp}: The baseline single-view method without reprojection loss, and novel view projected from the estimated altitude.
    \item \textbf{bp-tgt}: The baseline single-view method with the total loss in eq.~\ref{eq:loss_single}, with the novel view projected from the estimated altitude.
    \item \textbf{bp-ref}: The single-view method with the total loss in eq.~\ref{eq:loss_single}, with the novel view projected from the estimated altitude.
    \item \textbf{bp-tgt-lpips}:The same as \textbf{bp-tgt} but with LPIPs loss during training.
    \item \textbf{bp}: The single-view method without reprojection loss, and the novel view rendered from the warped frustum.
        \item \textbf{bf-tgt}: The baseline single-view method with the total loss in eq.~\ref{eq:loss_single}, with the novel view rendered from the warped frustum.
    \item \textbf{bf-ref}: The single-view method with the total loss in eq.~\ref{eq:loss_single}, with the novel view projected from estimated altitude.
    \item \textbf{bf-tgt-lpips}:The same as \textbf{bf} but with LPIPS loss during training.  
\end{itemize}
\textbf{bp} series denote the methods of directly obtaining the novel view image via projecting from the estimated altitude. \textbf{bf} series denote the methods of rendering the images from the warped frustum.
For altitude estimation, there are additional options for losses combinations:
\begin{itemize}
    \item \textbf{bl-pts}: The baseline method with the total loss in eq.~\ref{eq:loss_single} for a single view, with sparse altitude supervision from a randomly selected point cloud of size 100.
    \item \textbf{bf-pts}: The same loss setting with \textbf{bl-pts} but the novel view is synthesized from the warped frustum. 
\end{itemize}
\subsubsection{Novel view synthesis}
We compare the models quantitatively with the same image quality metrics in Table~\ref{tab:comp_nerf_mine}, the results are shown in Table~\ref{tab:single_nvs}, and the visualizations in Fig.~\ref{fig:single_nvs}. We test the models on the subset of the Masterpovitional3 (MP3) subset and compute the average values of all the metrics.
\begin{table}[H]
    \centering
     \caption{Impact of different losses for single-view rpcPRF on the MP3 subset SatMVS3DM Dataset.}    
    \begin{tabular}{c|c|cccc}
    \toprule
    Methods&Losses &\textbf{PSNR}$\uparrow$&\textbf{SSIM}$\uparrow$&\textbf{LPIPS}$\downarrow$ $\downarrow$ \\ 
    \midrule\multirow{4}{*}{projection}
    &\textbf{bp}&19.231&0.589&0.532\\
    &\textbf{bp-tgt}&22.619&	0.633&	0.301 \\
    &\textbf{bp-ref}&22.611 & 0.670& 0.248 \\
    &\textbf{bp-lpips}&23.486 & 0.473 & 0.485  \\
    \hline
    \midrule\multirow{4}{*}{frustum warping}
    &\textbf{bf}&20.620&0.547&0.489\\
    &\textbf{bf-tgt}& \textbf{25.499}& \textbf{0.806}& \textbf{0.194} \\
    &\textbf{bf-ref}&25.213 & 0.754& 0.201  \\
    &\textbf{bf-lpips}& 24.488 & 0.649& 0.596  \\
    \bottomrule
    \end{tabular}   
    \label{tab:single_nvs}
\end{table}
Table~\ref{tab:single_nvs} shows the crucial role of the reprojection loss in the single-view case. Both the supervisions of projections to the target view and the reference view take effects on improving the synthesized image quality on \textbf{bp} series and \textbf{bf} series. Besides, directly projecting the synthesized source view via RPC (\textbf{bp} series), yields worse results than \textbf{bf} series, and the best results are all achieved by  warping the whole frustum with supervision in eq.~\ref{eq:loss_single}, which shows the significance of frustum warping. Additionally, adding LPIPS supervision alongside reprojection loss during training guarantees sound image quality during training, but the generalization ability is weakened a lot.
\subsubsection{Altitude estimation}
\begin{figure*}
    \centering
    \includegraphics[width=18cm]{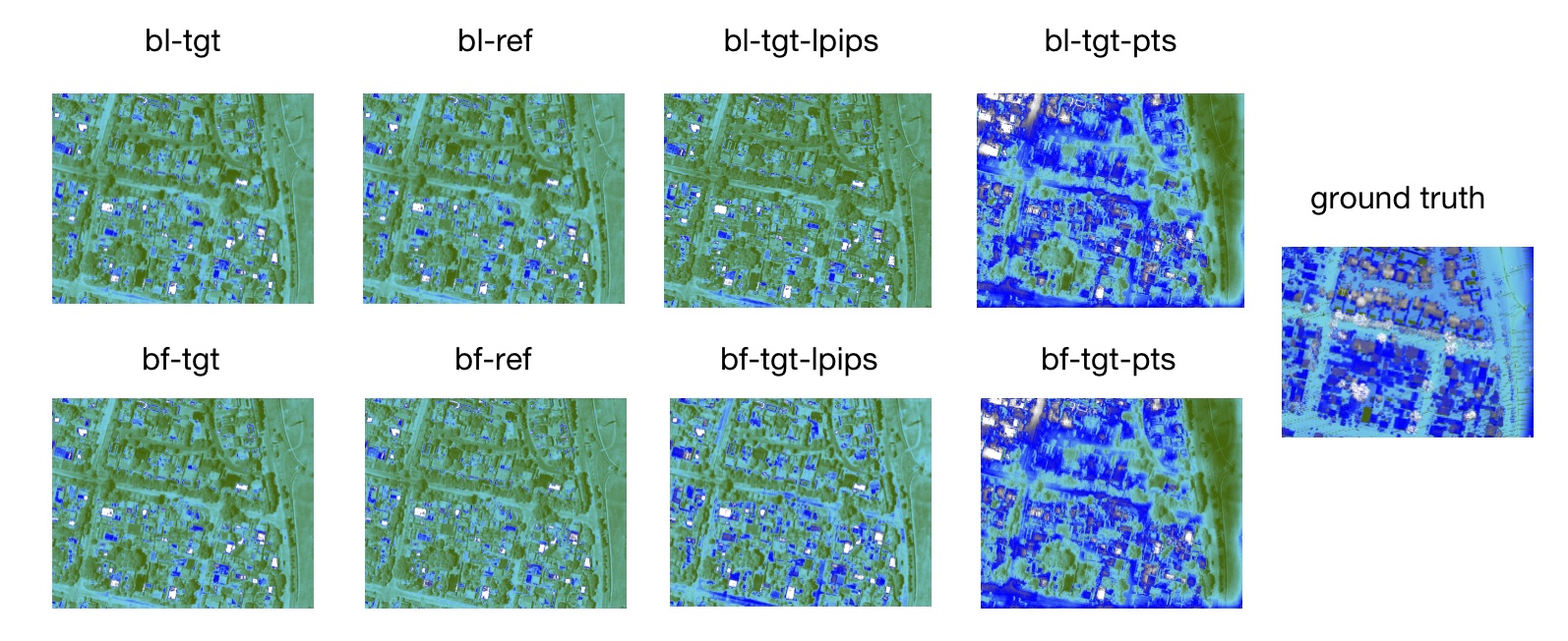}
    \caption{Qualitative comparison of different loss impacts on single-view rpcPRF for altitude estimation on the MP3 subset of Sat-MVS3DM Dataset. }
    \label{fig:single_altitude}
\end{figure*}

Unlike single-view learning settings with a scale factor and layered smooth loss \cite{li2021mine}, satellite photogrammetry does not appear in a layered fashion, and the depth range is larger a lot than the driving or indoor scenarios.
We evaluate single-view rpcPRF with different loss combinations for altitude estimation on   the subset MasterProvisional3 (MP3) from the Sat-MVS3DM dataset. The quantitative results are recorded in Table~\ref{tab:single_altitude} and the qualitative comparison is presented in Fig.~\ref{fig:single_altitude}. 
\begin{table*}
    \centering
     \caption{Single-view Altitude Estimation Results on TLC Dataset with different losses.}
      \resizebox{14cm}{!}{
    \begin{tabular}{c|c|cccccc}
    \toprule
    Methods&Losses &MAE(m)$\downarrow$& ME(m)$\downarrow$ & $<2.5$m(\%)$\uparrow$& $<5.0$m(\%)$\uparrow$& $<7.5$m(\%) $\uparrow$\\
    \midrule
    \multirow{4}{*}{projection}&\textbf{bp-tgt}&5.227&3.639&		19.1&61.4&78.4  \\ 
        &\textbf{bp-ref}& 5.238&3.654&	18.4&	61.0&78.3\\ 
    &\textbf{bp-tgt-lpips}&5.195&3.637&20.2 &61.9&79.9 \\
    &\textbf{bl-tgt-pts}&3.472&2.994 & 47.3&72.8&90.5  \\
    \hline
    \multirow{4}{*}{frustum warping}&
    \textbf{bf-tgt}& 5.197&	3.621&	22.4&	62.1&78.7\\
    &\textbf{bf-ref}& 5.202&3.592&21.9&70.4&81.5 \\
    &\textbf{bf-tgt-lpips}&4.843& 3.126&25.5&71.3&81.8 \\
    &\textbf{bf-tgt-pts}& \textbf{3.265}&	\textbf{2.725}&\textbf{51.8}&\textbf{74.9}&	\textbf{92.2}\\
    \bottomrule
    \end{tabular}
    }
    \label{tab:single_altitude}
\end{table*}
Table~\ref{tab:single_altitude} reveals the crucial role of sparse point supervision. 
%
The estimated altitude is the expectation value of the altitude at which the ray stops, which means that the accurate transparency map determines the real altitude. However, the transparency maps output by the model tend to reveal the best probability density to synthesize the RGB image, thus the shadows, the different textures patches in the RGB image may mislead the model for right transparency map. Sparse ground truth point cloud helps the model to regress to the correct transparency maps, with better reconstruction results. 

\subsection{Multiview rpcPRF and ablation study}
Similar as the single-view settings, this section studies on the impacts of the same two factors, the method for obtaining the novel view and the losses combinations.
\subsubsection{Model Notations}
\begin{itemize}
    \item \textbf{bp}: The baseline method with the total loss in eq.~\ref{eq:loss_mvs} for multiple view, with the novel view projected from estimated altitude.
    \item \textbf{bp-reproj}: The same as \textbf{bp}  with the supervision of $\mathcal{L}_{reproj}$.
    \item \textbf{bp-lpips}: The same as \textbf{bp} but with the LPIPS loss during training.
      \item \textbf{bf}: The baseline method with the total loss in eq.~\ref{eq:loss_mvs} for multiple views, and the novel view is synthesized from the warped frustum.
        \item \textbf{bf-reproj}: The same as \textbf{bf} but without the supervision of $\mathcal{L}_{reproj}$.
    \item \textbf{bf-lpips}: The same as \textbf{bf} but with LPIPS loss during training.   
\end{itemize}
Besides, the optional losses for multiview altitude estimation are again set to be:
\begin{itemize}
    \item \textbf{bl-pts}: The baseline method with the total loss in eq.~\ref{eq:loss_mvs} for multiview, and the novel view is projected from estimated altitude.
    \item \textbf{bf-pts}: The same loss setting with \textbf{bl-pts} but the novel view is synthesized from the warped frustum. 
\end{itemize}
\subsubsection{Novel view synthesis}
We compare the models quantitatively with the same image quality metrics in Table~\ref{tab:comp_nerf_mine}, the results are shown in Table~\ref{tab:comp_multi_bl}, and the visualizations are in Fig.~\ref{fig:comp_multi_bl}.
\begin{table}[H]
    \centering
     \caption{Impact of different losses for multiview rpcPRF on the MP3 subset SatMVS3DM Dataset.}
    \begin{tabular}{c|c|ccc}
    \toprule
    Methods &Losses&\textbf{PSNR}$\uparrow$&\textbf{SSIM}$\uparrow$&\textbf{LPIPS}$\downarrow$ \\ 
    \midrule\multirow{3}{*}{projection}
    &\textbf{bp}& 21.037& 0.528& 0.436  \\
    &\textbf{bp-reproj}& 20.942&0.587&0.430   \\
    &\textbf{bp-lpips}&23.325&0.690&	0.196  \\
    \hline
    \multirow{3}{*}{frustum warping}
    &\textbf{bf}& 24.367&0.710&0.207 \\
    &\textbf{bf-reproj}&\textbf{26.323}&0.801&0.199\\
    &\textbf{bf-lpips}&24.262&	\textbf{0.827}&\textbf{0.122}\\
    \bottomrule
    \end{tabular}   
    \label{tab:comp_multi_bl}
\end{table}
As shown in Table~\ref{tab:comp_multi_bl}, rendering a novel view from the warped frustum still takes the lead, and the self-reprojection loss helps rpcPRF to achieve the highest PSNR, while increasing SSIM and decreasing LPIPS. LPIPS supervision also boosts multiview rpcPRF performance, but reprojection loss enables a more comprehensive enhancement. 
\subsubsection{Altitude estimation}
\begin{figure*}
        \centering
    \includegraphics[width=18cm]{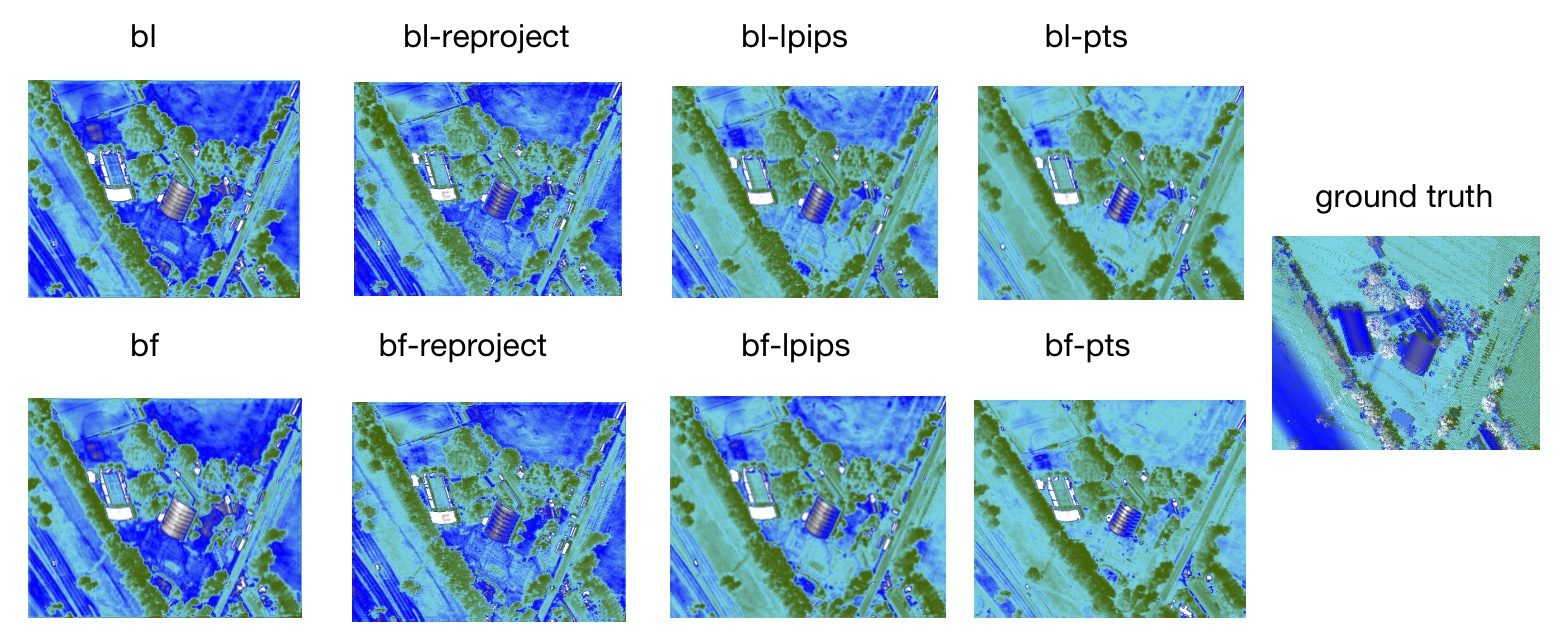}
    \caption{Qualitative comparison of different loss impacts on multiview rpcPRF for altitude estimation on the MP3 subset of Sat-MVS3DM Dataset. }
    \label{fig:multi_altitude}
\end{figure*}

We list the quantitative results in Table~\ref{tab:multi_altitude} and qualitative results in Fig.~\ref{fig:multi_altitude}. The sparse point supervision again helps the multiview version of rpcPRF to achieve more accurate altitude estimation for all the metrics in Table~\ref{tab:multi_altitude}.
\begin{table*}
    \centering
     \caption{Multiview altitude estimation results Comparison with different losses.}
     \resizebox{14cm}{!}{
    \begin{tabular}{c|c|ccccc}
    \toprule
   Methods&Losses&MAE(m)$\downarrow$& ME(m)$\downarrow$ & $<2.5$m(\%)$\uparrow$& $<5.0$m(\%)$\uparrow$& $<7.5$m(\%) $\uparrow$ \\
    \midrule
    \multirow{4}{*}{projection} &
    \textbf{bp}&5.521&	4.386& 19.5&66.0&76.4 \\    
    &\textbf{bp-reproj}&5.477&4.313&19.6&67.6&76.6  \\
    &\textbf{bp-lpips}&5.181&	3.617&	20.4&	62.6&78.9 \\
    &\textbf{bl-pts}& 3.901&2.571&	48.2&	76.9&86.7 \\
        \hline
        \multirow{5}{*}{frustum warping}
    &\textbf{bf}&4.843&	3.403&20.5&	70.4&81.8 \\
    &\textbf{bf-reproj}&4.858&3.426&20.5&69.8& 81.7\\
    &\textbf{bf-lpips}& 4.869& 3.421&20.5&69.8& 81.9 \\
    &\textbf{bf-pts}& \textbf{3.223}&\textbf{2.661}&\textbf{45.0}&	\textbf{79.8}&\textbf{91.1}  \\
    &\textbf{bf-lpips-pts}&3.888&2.879&43.4&72.8&86.4\\  
    \bottomrule
    \end{tabular}
   }
    \label{tab:multi_altitude}
\end{table*}

\subsection{Investigation on the number of views}
 This part studies the effect of Input Number of views.  
 As shown in Table~\ref{tab:abla_Nview}, the synthesized image quality increases slightly with more of input views, according to the three metrics. In most cases, single and triple inputs are enough, for overall consideration.
\begin{table}[H]
    \centering
       \caption{Comparison of model performance on novel view synthesis metrics,  memory usage, parameter amount, and inference time with different altitude sample numbers.} 

    \begin{tabular}{c|ccc}
    \toprule
        $N_v$& \textbf{PSNR}$\uparrow$&\textbf{SSIM}$\uparrow$ &\textbf{LPIPS}$\downarrow$\\
        \midrule
        1& 25.499& 0.806& 0.281 \\
         3& 26.323& 0.801& 0.199 \\
         5& 26.889&	0.832&	0.152 \\
         7& 27.037&0.868&0.148 \\
         9&\textbf{27.982} &\textbf{0.887}&\textbf{0.141} \\
         \bottomrule
         
    \end{tabular}
    \label{tab:abla_Nview}
\end{table}
\subsection{Investigation on the number of altitude samples}
This part analyzes the sensitivity of rpcPRF to the altitude sample frequency of the MPI. To construct the MPI with the altitude decoder, we sample altitude of size 16, 24, 32, 48 respectively to feed the decoder and to construct the warping grids. Therein the decoder yields MPI with corresponding sizes, and we quantitatively compare the effect on SatMVS3DM dataset for Novel Synthesis in Table~\ref{tab:abla_Nsample}.
\begin{table}
    \centering
        \caption{Comparison of model performance on novel view synthesis metrics,  memory usage, parameter amount, and inference time with different altitude sample numbers.}
\resizebox{9cm}{!}{
    \begin{tabular}{c|ccc|ccc}
    \toprule
        $N_s$& \textbf{PSNR}$\uparrow$&\textbf{SSIM}$\uparrow$&\textbf{LPIPS}$\downarrow$&\textbf{Memory}$\downarrow$&\textbf{Params}$\downarrow$& \textbf{FLOPs}$\downarrow$ \\
        \midrule
         16& 24.356&0.728&0.336&\textbf{0.319G}&19.79M&\textbf{104.95G} \\
         24& 25.499&0.752&0.259&0.397G&19.79M&153.59G \\
         32& 25.499& \textbf{0.806} &\textbf{0.194}&0.477G&19.79M&202.23G \\
         40&\textbf{27.368} &0.731&0.202&0.554G&19.79M&250.87G \\
         \bottomrule
    \end{tabular} }
    \label{tab:abla_Nsample}
\end{table}

Apparently, the memory consumption and model parameters increase rapidly with the number of samples, and the corresponding FLOPs grows the fastest especially.
Meanwhile the synthesized image quality has no radical changes with the altitude samples.
\subsection{Investigation on the number of sparse points as supervision}
This part investigates the number of sparse points used as supervision for the single-view altitude estimation tasks. Except for the number of points, all the compared models follow the setting of \textbf{bf-pts}. Table~\ref{tab:abla_Npoints} shows that the performance of altitude estimation improves when the number of points for supervision is relatively small, but too many points negatively affect the model performance.
\begin{table}[H]
    \centering
     \caption{Single-view altitude estimation results with varying number of sparse points $N_{pts}$ as supervision.}
     \resizebox{9cm}{!}{
    \begin{tabular}{c|cccccc}
    \toprule
    $N_{pts}$ &MAE(m)$\downarrow$& ME(m)$\downarrow$ & $<2.5$m(\%)$\uparrow$& $<5.0$m(\%)$\uparrow$& $<7.5$m(\%) $\uparrow$\\
    \midrule
    0& 5.202& 3.592 &21.9 &70.4 &81.5\\
    20& 4.649&5.063&27.5&49.5&75.3 \\
    50& 3.814&2.825&42.5&73.8&88.4\\
    100& \textbf{3.265}&	\textbf{2.725}&\textbf{51.8}&\textbf{74.9}&	\textbf{92.2}\\
    200& 4.174&3.124&33.5&71.6&81.3\\
    \bottomrule
    \end{tabular}   }
    \label{tab:abla_Npoints}
\end{table}

\section{Conclusion}
In this paper, we propose rpcPRF, a generalization of MPI and NeRF models for rational polynomial camera photogrammetry, with single-view and multiple view versions. 
Given a single image or a triple of images, rpcPRF synthesize novel image from a novel rpc for the new view. 
rpcPRF is the first work to use MPI for pushbroom sensors, for which we introduce the frustum warping module.
Other than training and testing within one scene, we propose to use self-reprojection supervision, thus rpcPRF is able to synthesize a novel view for an entirely new scene. 
We conduct extensive evaluations on the single-view and multiview versions of rpcPRF.
The proposed rpcPRF shows competence in synthesized image quality, with much faster inference speed and lower computation costs. We also demonstrate by the experiments the effectiveness of the proposed frustum warping module.
Moreover, rpcPRF is able to synthesize altitude maps without dense depth supervision. With sparse altitude clues, both single-view and multiview rpcPRF series  can reconstruct the altitude map.
In the future, we will further explore strategies to reduce the training memory caused by intermediate computation of MPI at different scales. The efficiency of inferring the ray termination is also a pivotal topic for further research, for which hierarchical sampling on MPI might be a good choice.

\bibliographystyle{IEEEtran}
\bibliography{ref}

\end{document}